\documentclass[final,5p,times,twocolumn]{elsarticle}
\usepackage{amssymb}
\usepackage{amsmath,amssymb,amsfonts}
\usepackage{algorithmic}
\usepackage{graphicx}
\usepackage{textcomp}

\usepackage{booktabs}
\usepackage{threeparttable}
\usepackage{multirow}
\usepackage{bigstrut}
\usepackage{subfigure}
\usepackage{array}
\usepackage{algorithm}
\usepackage{enumerate}
\usepackage{epstopdf}
\usepackage{bm}
\usepackage[nodots]{numcompress}

\usepackage{float}
\usepackage{fancyhdr}
\usepackage{threeparttable}
\usepackage{ragged2e}
\usepackage{url}
\usepackage[table]{xcolor}
\usepackage{stfloats}

\usepackage{etoolbox}
\biboptions{sort&compress}
\usepackage{xpatch}
\makeatletter
\def\changeBibColor#1{%
  \in@{#1}{}
  \ifin@\color{blue}\else\normalcolor\fi
}
\xpatchcmd\@bibitem
  {\item}
  {\changeBibColor{#1}\item}
  {}{\fail}

\xpatchcmd\@lbibitem
  {\item}
  {\changeBibColor{#2}\item}
  {}{\fail}
\makeatother

\newcommand{\tabincell}[2]{\begin{tabular}{@{}#1@{}}#2\end{tabular}}

\begin{document}

\begin{frontmatter}

\title{NaturalAE: Natural and Robust Physical Adversarial Examples for Object Detectors}

\author[a]{Mingfu~Xue}
\ead{mingfu.xue@nuaa.edu.cn}
\author[a]{Chengxiang~Yuan}
\ead{yuancx@nuaa.edu.cn}
\author[a]{Can He}
\ead{hecan@nuaa.edu.cn}
\author[a]{Jian Wang}
\ead{wangjian@nuaa.edu.cn}
\author[b]{Weiqiang Liu}
\ead{liuweiqiang@nuaa.edu.cn}
\address[a]{College of Computer Science and Technology, Nanjing University of Aeronautics and Astronautics, Nanjing, China}
\address[b]{College of Electronic and Information Engineering, Nanjing University of Aeronautics and Astronautics, Nanjing, China}

\begin{abstract}
Recently, many studies show that deep neural networks (DNNs) are susceptible to adversarial examples, which are generated by adding imperceptible perturbations to the input of DNN.
However, in order to convince that adversarial examples are real threats in real physical world, it is necessary to study and evaluate the adversarial examples in real-world scenarios.
In this paper, we propose a natural and robust physical adversarial example attack method targeting object detectors under real-world conditions, which is more challenging than targeting image classifiers. The generated adversarial examples are robust to various physical constraints and visually look similar to the original images, thus these adversarial examples are natural to humans and will not cause any suspicions.
First, to ensure the robustness of the adversarial examples in real-world conditions, the proposed method exploits different image transformation functions (Distance, Angle, Illumination and Photographing), to simulate various physical changes during the iterative optimization of the adversarial examples generation.
Second, to construct natural adversarial examples, the proposed method uses an adaptive mask to constrain the area and intensities of the added perturbations, and utilizes the real-world perturbation score ($RPS$) to make the perturbations be similar to those real noises in physical world.
Compared with existing studies, our generated adversarial examples can achieve a high success rate with less conspicuous perturbations.
Experimental results demonstrate that, the generated adversarial examples are robust under various indoor and outdoor physical conditions, including different distances, angles, illuminations, and photographing. Specifically, the attack success rate of generated adversarial examples indoors and outdoors is high up to 73.33\% and 82.22\%, respectively.
Meanwhile, the proposed method ensures the naturalness of the generated adversarial example, and the size of added perturbations is as low as 29361.86, which is much smaller than the perturbations in the existing works (95381.14 at the highest).
Further, the proposed physical adversarial attack method can be transferred from the white-box models to other object detection models. The attack success rate of the adversarial examples (generated targeting Faster R-CNN Inception v2) is high up to 57.78\% on the SSD models, while the success rate of adversarial example (generated targeting YOLO v2) on SSD models reaches 77.78\%.
This paper reveals that physical adversarial example attacks are real threats in the real-world conditions, and can hopefully provide guidance for designing robust object detectors and image classifiers.
\end{abstract}


\end{frontmatter}

\section{Introduction}
\label{intro}
In recent years, deep neural networks (DNNs) have made significant breakthroughs and are widely applied in many areas, \textit{e.g.}, image classification \cite{krizhevsky2012imagenet,he2016deep}, speech recognition \cite{greff2016lstm,hinton2012deep}, natural language processing \cite{collobert2008unified,kumar2016ask}, self-driving cars \cite{bojarski2016end,chen2015deepdriving} and smart healthcare \cite{faust2018deep,ravi2016deep}.
However, many recent researches indicate that DNNs are susceptible to adversarial examples \cite{szegedy2013intriguing, goodfellow2014explaining, kurakin2016adversarial, carlini2017towards, papernot2016transferability}, where the attackers can craft well-designed inputs to cause the target machine learning models to output incorrect predictions \cite{zhang2018adversarial}. As a result, the adversarial examples can bring serious consequences to security and safety critical systems \cite{yuan2019adversarial}, such as autonomous vehicles and face recognition systems.

In the literature, many digital adversarial example generation methods have been proposed, such as the fast gradient sign method (FGSM) \cite{goodfellow2014explaining}, the basic iterative method (BIM) \cite{kurakin2016adversarial}, the momentum iterative method (MIM) \cite{dong2017boosting} and the Carlini-Wagner (C\&W) method \cite{carlini2017towards}.
In addition, a few studies \cite{athalye2018synthesizing, eykholt2018robust, jan2019connecting} have indicated that, the DNN models are vulnerable to the adversarial examples in real physical world.
Unlike the digital adversarial example attacks, the physical adversarial examples need to adapt to different physical conditions, such as different distances, angles and illuminations, which makes the attacks more challenging.
Athalye \textit{et al.} \cite{athalye2018synthesizing} propose the Expectation Over Transformation (EOT) algorithm to improve the physical robustness of the generated adversarial examples. Eykholt \textit{et al.} \cite{eykholt2018robust} add perturbations to a set of clean images collected in different physical conditions. Then, the objective function is optimized based on these images to obtain adversarial examples.
Jan \textit{et al.} \cite{jan2019connecting} use a generative adversarial networks to simulate the physical conditions. The targets of the above adversarial attacks are the image classifiers.

A few works \cite{lu2017adversarial,chen2018shapeshifter, song2018physical, zhao2018practical} propose adversarial attacks targeting the object detectors, which are more difficult than attacking the image classifiers. The reason is that, the image classifier can only classify an image into a single class, but the object detector needs to classify multiple objects in an image and determine the position of each object.
For example, the work \cite{lu2017adversarial} captures a set of pictures that contain the target object (\textit{i.e.}, stop sign) as the training data, and minimizes the confidence score of stop sign in all region proposals to optimize the generated adversarial perturbations.
However, the method proposed in \cite{lu2017adversarial} requires to add very large perturbations to their generated adversarial examples to fool the detector, and people even cannot distinguish the original objects.
Chen et al. \cite{chen2018shapeshifter} exploit the Expectation Over Transformation (EOT) \cite{athalye2018synthesizing} technique to attack the object detector, and perform two targeted adversarial attacks (target classes are ``person'' and ``sports ball'') and one untargeted adversarial attack against the Faster Regions with Convolutional Neural Networks (Faster R-CNN) detector.
In \cite{chen2018shapeshifter}, the difference between the generated perturbations and the background of original image is conspicuous, thus humans can easily perceive the anomalies in the adversarial examples.

In this paper, for the first time, we propose a natural and robust physical adversarial example generation method against the object detectors.
Specifically, for robustness, the proposed method performs a series of transformations to simulate the physical conditions during the optimization process of generating an adversarial example.
Besides, an adaptive mask is used to constrain the size of added perturbations, and the real-world perturbation score ($RPS$) is proposed to optimize the perturbations at each iteration, so that the generated adversarial examples look more natural.
In our experiments, we print our generated adversarial examples and those ones that generated in works \cite{lu2017adversarial,chen2018shapeshifter}.
We take photos for these printed images in real physical world, and submit the photos to Faster R-CNN Inception v2 \cite{ren2016faster} and YOLO (You Only Look Once) v2 \cite{redmon2017yolo9000} models to evaluate the performances of the adversarial examples under indoor and outdoor physical scenes.
The experimental results indicate that, compared with these existing methods \cite{lu2017adversarial,chen2018shapeshifter}, the proposed method can ensure the visual naturalness of generated adversarial examples, while the physical attack performances of proposed method are close to the state-of-the-art works.

The contributions of this work are as follows:
\begin{itemize}
  \item {\textbf{Robustness.}
    We propose a physical adversarial examples generation method for object detectors, which is more challenging than targeting the image classifiers. We perform various image transformations for a generated adversarial example to ensure its robustness when facing various practical conditions, including different angles, distances, illuminations and photographing. Experimental results show that, the attack success rate of generated adversarial examples indoors and outdoors is high up to 73.33\% and 82.22\% respectively, which is better or close to that of the adversarial examples generated by existing methods \cite{lu2017adversarial,chen2018shapeshifter}.}
  \item {\textbf{Naturalness.}
    We propose two novel techniques (adaptive mask and $RPS$) to constrain and optimize the added adversarial perturbations, so as to make the generated adversarial example look more natural.
    The adaptive mask can limit the area and intensities of the added perturbations, while the $RPS$ can make the perturbations be close to real-world noises.
    In this way, the generated adversarial example is more natural and looks similar to the aging version (\textit{i.e.}, eroded by rain and sunlight) of the original image.
    Compared with the existing works \cite{lu2017adversarial,chen2018shapeshifter}, the proposed method can obtain a good balance between the robustness and naturalness of a generated adversarial example.
    In other words, the generated adversarial example can achieve a high success rate with less conspicuous perturbations, which will not cause human's suspicions.
  }
  \item {\textbf{Transferability.} The adversarial examples (Proposed-1, Proposed-2 and Proposed-3) generated by the proposed method are demonstrated to be transferable. Experimental results show that, the adversarial examples generated targeting Faster R-CNN Inception v2 \cite{ren2016faster} and YOLO v2 \cite{redmon2017yolo9000} models can be successfully transferred to different Single Shot Detector (SSD) models (SSD Inception v2 \cite{Liu2016SSD} and SSD MobileNet v2 \cite{Liu2016SSD}). The transfer success rates of Proposed-1 and Proposed-2 on SSD models are high up to 51.11\% and 57.78\%, while the transfer success rate of Proposed-3 on SSD models even reaches 77.78\%. This demonstrates that, the proposed method can be applied to attack different object detection models which have completely different working mechanisms and network structures from the white-box detectors.
  }
\end{itemize}

This paper is organized as follows. Section \ref{sec:preliminary} reviews current deep learning based object detectors and related adversarial example attack methods. The threat model of the adversarial attack method is described in Section \ref{sec:threat_model}. Section \ref{sec:proposed_method} elaborates the proposed adversarial example generation method. Experimental evaluations in the physical conditions are presented in Section \ref{sec:experiment}. Section \ref{sec:conclusion} concludes this paper.

\section{Preliminary}\label{sec:preliminary}

We first overview the deep learning based object detectors, which are the targets of the adversarial attacks in this work. Then, we review related adversarial example attack methods from two aspects: digital adversarial example attacks and physical adversarial example attacks.

\subsection{Object detectors}

Image classification and object detection are two basic tasks for machine learning models.
The image classification task focuses on the overall content of an image, which aims to classify an input image into a single class.
However, the object detector recognizes all the possible objects in an image, which can determine the category and location for each object.
There are two different detection strategies for the current object detectors based on deep learning: 1) two-step detection strategy, such as R-CNN \cite{girshick2014rich}, Fast R-CNN \cite{girshick2015fast} and Faster R-CNN \cite{ren2016faster}; 2) and one-step detection strategy, such as SSD \cite{Liu2016SSD}, YOLO \cite{redmon2016you}, YOLO v2 \cite{redmon2017yolo9000} and YOLO v3 \cite{redmon2018yolov3}. The two-step object detectors first generate region proposals that may contain objects in an image, then classify the objects in each region proposal. The one-step object detectors directly predict the class and position of each object in an image by running a single Convolutional Neural Network (CNN).

\subsection{Adversarial example attack methods}
Given a trained machine learning model $f(\cdot)$ and an input $x$, the model will output a predicted label. Ideally, the predicted output should be the true label $y$. The adversarial attack needs to find an input $x'$ that is similar to the original input $x$, but this input $x'$ can cause the model to produce an erroneous output \cite{carlini2017towards}. Generally, adversarial attacks can be divided into untargeted attacks and targeted attacks. The untargeted attack only requires to make the model output a wrong prediction \cite{dong2017boosting}, \textit{i.e.}, be different from the true label $y$, while the targeted attack causes the model to produce a specific label $ y'$ $(y' \neq y)$.
Early adversarial example attacks \cite{goodfellow2014explaining,kurakin2016adversarial,carlini2017towards,dong2017boosting} focus on generating digital adversarial examples, which can achieve high success rates under the laboratory/digital settings.
To convince that adversarial examples are real threats in practice, some works \cite{athalye2018synthesizing, eykholt2018robust, jan2019connecting,lu2017adversarial, song2018physical, zhao2018practical, chen2018shapeshifter} have proposed and evaluated the adversarial example attacks in physical conditions.
However, limited to various physical constraints (\textit{e.g.} distance, angle, illumination and photographing), those adversarial examples generated in digital domain have failed in the real world, or only achieved a low success rate \cite{lu2017no}.
Since the principles of the untargeted attack and the targeted attack are similar, we will take the untargeted attack as an example to review related digital adversarial attack methods and physical adversarial attack methods.

\textbf{Digital adversarial attacks:} The digital adversarial example generation methods mainly include gradient-based methods and optimization-based methods. The gradient-based methods add perturbations to the original image, which are obtained by computing the gradient of the loss function \cite{goodfellow2014explaining}. For example, in the FGSM method, the adversarial example $x'$ is calculated as follows \cite{goodfellow2014explaining}:
\begin{equation}
  x' = x + \epsilon  \cdot sign({\nabla _x}J_{f}(x,y))
\end{equation}
where the loss function $ J_{f}(\cdot)$ calculates the difference between the predicted label of the model and the true label $y$. ${\nabla _x}J_{f}(x,y)$ is the gradient of the loss function, and $\epsilon$ is a hyper-parameter. This one-step gradient-based method is fast, but it is difficult to generate adversarial examples with high success rate by adding perturbations only once. Hence, many iterative gradient-based methods are proposed to generate adversarial examples. The basic iterative method for adversarial examples generation is as follow \cite{kurakin2016adversarial}:
\begin{equation}
    x_{0}^{'} = x,~~~ x_{m+1}^{'} = x_{m}^{'} + \lambda  \cdot sign({\nabla _x}J_{f}(x_{m}^{'},y))
\end{equation}
where $m$ is the number of iterative optimization, and $\lambda $ is a hyper-parameter.

On the other hand, the optimization-based methods formulate the adversarial examples generation as an optimization problem. For instance, the Carlini-Wagner attack method \cite{carlini2017towards} generates adversarial examples by minimizing the following object function \cite{carlini2017towards}:
\begin{equation}
   \mathop {\arg \min }\limits_\delta  {\rm{  }}\alpha {\left\| \delta  \right\|_p} - J_{f}(x + \delta ,y)
\end{equation}
where $ \left \| \delta \right \|_{p} $ is the $l_{p}$ norm of the added perturbations $\delta$, and $\alpha$ is a parameter used to adjust the weight of perturbations.

\textbf{Physical adversarial attacks:}
The physical adversarial attack aims to fool the target model (image classifiers or object detectors) that deployed in the real physical world.
The adversarial attacks on image classifiers only require to make the model misclassify an image. However, the adversarial attacks on object detectors need to cause the target model misclassify the objects in each detected bounding box, which are more difficult than the adversarial attacks on image classifiers.
In \cite{athalye2018synthesizing, eykholt2018robust, jan2019connecting}, the authors generate physical adversarial examples for image classifiers,
while in \cite{lu2017adversarial, song2018physical, zhao2018practical, chen2018shapeshifter}, the authors generate physical adversarial examples for object detectors.

Taking the methods in \cite{eykholt2018robust} and \cite{zhao2018practical} as examples, we describe the adversarial examples generation methods for image classifiers and for object detectors in the physical conditions, respectively.
In \cite{eykholt2018robust}, the optimization method is used to search for the perturbations $\delta$ that can fool the image classifier.
To ensure the physical robustness of the generated adversarial examples, the method uses real-world images to optimize the added perturbations in the process of generating adversarial examples.
The method generates adversarial examples by solving the following objective function \cite{eykholt2018robust}:
\begin{equation}
   \mathop {\arg \min }\limits_\delta  {\rm{  }}\alpha {\left\| \delta  \right\|_p} - {1 \over k}\sum\limits_{i = 1}^k{J_{f}(x_{i} + \delta ,y)}
\end{equation}
where $x_{i}$ is an original image, and $k$ is the number of the original images.

The main differences between the work \cite{eykholt2018robust} and this paper are as follows.
First, the work \cite{eykholt2018robust} targets at the image classifiers, while this paper launches the physical adversarial example attacks against the object detectors. As discussed in Section \ref{intro}, it is more difficult to launch the adversarial attacks against the object detectors.
Second, the work \cite{eykholt2018robust} crafts the stickers and pastes them on the stop sign to launch the physical attacks, or print the posters to launch the attacks.
However, the crafted stickers/posters in \cite{eykholt2018robust} are obvious and look unnatural, which will cause the failure of adversarial attacks.
In this paper, we propose two novel techniques ($RPS$ and adaptive mask) to ensure the naturalness of our adversarial examples. Specifically, we generate the adversarial road signs and print them to launch the adversarial example attacks in real physical world.
Meanwhile, the physical attack performances of the proposed method are better or close to the existing works \cite{lu2017adversarial, chen2018shapeshifter}.
Finally, the work \cite{eykholt2018robust} proposes the $RP2$ (Robust Physical Perturbations) method to construct the adversarial stickers/posters, while the proposed method exploits the $EOT$ (Expectation Over Transformation \cite{athalye2018synthesizing}) attack framework to generate the adversarial examples, which can better simulate the physical constraints in real physical world.

Zhao \textit{et al.} \cite{zhao2018practical} proposes two types of physical adversarial example attacks, hiding attack (HA) and appearing attack (AA).
The hiding attack aims to cause the detector fail to recognize the target objects \cite{zhao2018practical}, while appearing attack aims to cause the detector incorrectly recognize the generated adversarial examples as the specific objects.
Huang \textit{et al.} \cite{huang2019upc} develop an Universal Physical Camouflage (UPC) attack for object detectors, which can attack all instances of the same target class (\textit{e.g.}, all cars in an input image) with the generated universal pattern.

The above attack methods \cite{lu2017adversarial, song2018physical, zhao2018practical, chen2018shapeshifter, huang2019upc} add the perturbations into the target objects to generate the adversarial examples, while some other works \cite{HuangKL19,Lee2019On,LiSK19Adversarial} can launch the physical adversarial attacks without manipulating the target objects.
Huang \textit{et al.} \cite{HuangKL19} craft an adversarial example that looks like the advertising signboard. By placing the adversarial signboard below the target objects with a certain distance, the Faster R-CNN detector fails to detect the target objects (\textit{e.g.}, the stop signs).
Lee \textit{et al.} \cite{Lee2019On} generate a physical adversarial patch, which can prevent the YOLO v3 from detecting any objects in a submitted image.
Li \textit{et al.} \cite{LiSK19Adversarial} directly manipulate the camera device and constructs a translucent adversarial sticker. They paste the sticker on the lens of a camera and attack the DNN based classifiers.

However, these existing studies \cite{zhao2018practical, huang2019upc} add obvious perturbations on the target objects, which cause their generated adversarial examples can be easily perceived by humans.
Although the recent works \cite{HuangKL19,Lee2019On,LiSK19Adversarial} do not manipulate the target objects, their generated adversarial signboard \cite{HuangKL19} and patch \cite{Lee2019On} are very large, which make them look strange compared to the surrounding environments. In addition, the adversarial sticker attack \cite{LiSK19Adversarial} against the camera is difficult to be conducted in real-world conditions, as the sticker can be easily observed by humans when it is pasted on the lens of the camera.
Compared with these existing methods, the proposed method aims to achieve a good balance between the robustness and the naturalness of a generated adversarial example. First, we perform a series of image transformations to simulate different physical constraints during the iterative optimization, so as to ensure the robustness of our generated adversarial examples.
Second, we propose two novel techniques, the adaptive mask and real-world perturbation score, to constrain the size (i.e., area and intensities) of the added perturbations and make the perturbations look like real-world noises, respectively.
As a result, our generated adversarial examples can achieve robust attack performance with more natural perturbations.

\begin{figure*}[htbp]
  \centering
    \includegraphics[width=4.8in]{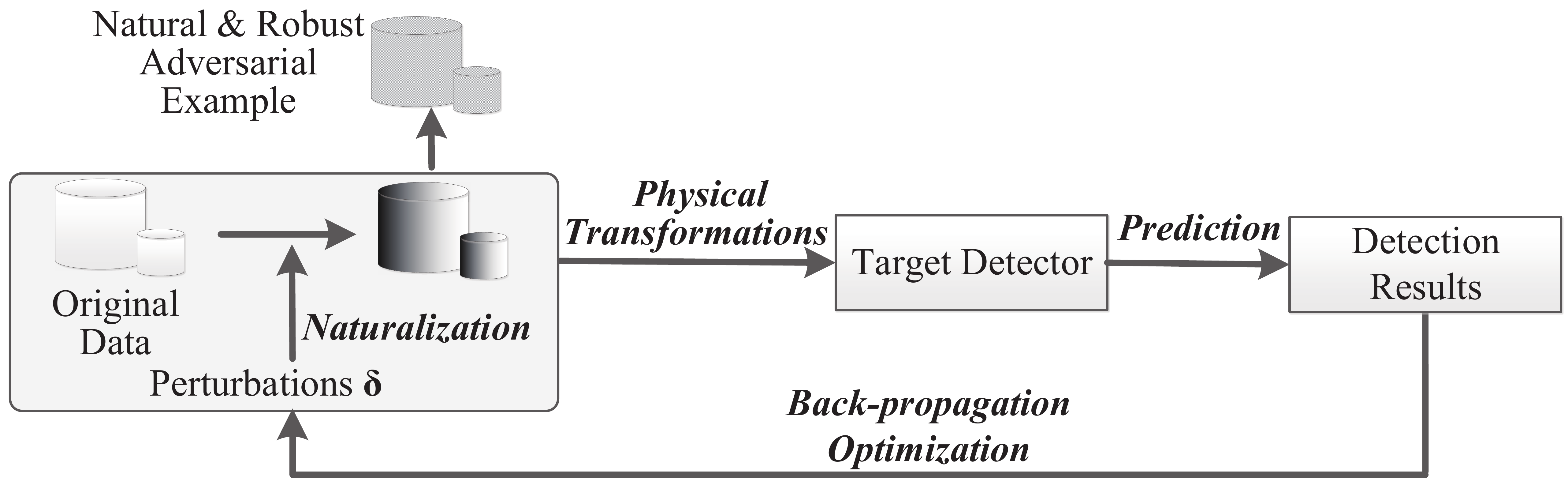}
  \caption{Overall flow of the proposed adversarial examples generation method.}
  \label{fig:overall}
\end{figure*}

\section{Threat model}\label{sec:threat_model}

In this section, we describe the potential adversarial attackers in terms of the adversary's goals and adversary's abilities. Specifically, to ensure the generated adversarial examples more robust and visually more natural, we make some additional constraints on the added perturbations.

\subsection{Adversary's goals}
This work aims to launch the physical adversarial example attacks against the object detectors.
The goal of an physical adversarial example attacker is to fool the target object detector to make incorrect predictions.
Compared with the attacks on image classifiers, it is more challenging to generate adversarial examples to attack the object detectors.
In this paper, we select two state-of-the-art object detectors (Faster R-CNN Inception v2 \cite{ren2016faster} and YOLO v2 \cite{redmon2017yolo9000} models) as the targets of our proposed adversarial attacks.
These two detectors can detect the objects in an image with a high detection accuracy and a fast detection speed \cite{ren2016faster,redmon2017yolo9000}.
Formally, given an input $x$, the detection model $f(\cdot)$, and a true label $y$, the object detected on the bounding box $r_{i}$ $(r_{i} \in rpn(x))$ is $f(x_{r_{i}}) = y $.
The $rpn(x) = \{ {r_1},{r_2},...,{r_n}\} $ represents the collection of bounding boxes detected by the model, where $r_{i}$ is the $i$-th bounding box and $n$ is the number of all the bounding boxes.
The goal of the adversary is to add perturbations $\delta$ that satisfies the following rule to the original image:
\begin{equation}
  f(x_{r_{i}} + \delta ) \neq y
\end{equation}
To ensure that the generated adversarial examples are more natural and more robust in real-world conditions, we set the following restrictions on the added perturbations:
\begin{itemize}
  \item { The added perturbations should not affect humans' understanding of the object in the original image.}
  \item { The added perturbations should not be too anomalous and should be similar to real-world noises. In this way, the generated adversarial examples are more natural and will not arouse human's suspicions.}
  \item {The added perturbations should be robust in various physical conditions (\textit{e.g.}, different distances, angles, illuminations, photographing). In other words, the generated adversarial examples can fool the object detectors in different physical conditions.}
\end{itemize}

\subsection{Adversary's abilities}
Generally, according to the adversary's abilities, adversarial attacks can be categorized into white-box attacks, black-box attacks and gray-box attacks. In this work, adversaries generate adversarial examples in a white-box scenario. It means that adversaries have the knowledge of the target model, which includes model architectures, parameters, and weights. Although the proposed method generates adversarial examples on a white-box model, experimental results show that the adversarial examples generated by the proposed method can generalize well to other black-box models successfully. Therefore, by utilizing the transferability of the generated adversarial examples, adversaries can also carry out black-box attacks.

In real physical world, the inputs of a DNN based object detector is collected by its external camera, and an attacker cannot directly submit the generated adversarial examples to its internal DNN model.
Therefore, in this paper, we assume an adversary can only print the generated adversarial examples, and the image that captured by a camera will be used as the input of the target object detectors.

\section{The proposed attack method}\label{sec:proposed_method}

\subsection{Overall procedure}
Fig. \ref{fig:overall} presents the overall flow of the proposed adversarial examples generation method.
To illustrate that the proposed method can be applied to other fields, we describe the working procedure with a high-level schematic representation, rather than limiting to the road signs.
The proposed method works as follows.
First, the constrained perturbations are added to the original data to generate adversarial example.
Second, various physical transformations are applied to the generated adversarial example, so as to ensure the robustness in real physical world.
Then, the transformed data is used as the input of the object detector, and the adversarial perturbations will be optimized by calculating the gradient of the objective function.
The above iterative optimization steps are repeated until the value of the objective function is less than a predefined threshold.
Finally, the natural and robust physical adversarial examples will be generated.

In this paper, we target at the road sign and generate adversarial examples to attack the DNN based object detectors.
Different image transformation functions are introduced in Section \ref{sec:proposed_method_1}.
The methods of constraining the added perturbations are introduced in Section \ref{sec:proposed_method_2}.

\subsection{Image transformations for simulating different physical conditions} \label{sec:proposed_method_1}
In this paper, to ensure the robustness of the generated adversarial examples in the real world, we apply a series of image transformations to these modified images.
Inspired by the existing work \cite{athalye2018synthesizing}, we adopt the \textit{Expectation Over Transformation (EOT)} technique \cite{athalye2018synthesizing} to perform the transformations.
Such image transformations simulate the images in different physical conditions, including distance, angle, illumination and photographing.
Specifically, at each iteration of the optimization process, the above image transformations will be performed on the generated adversarial example, to make the added perturbations more robust.

The transformation function is denoted as $t(\cdot)$. In this paper, we considers the following four different image transformation methods.
\begin{itemize}
  \item {\textbf{Distance transformation}: The distance transformation function first randomly adjusts the size of the image. Then, according to the size of the image, different size of Gaussian kernels are used to transform the image. In real physical world, if the distance between an object and the detector is longer, the object would be smaller and looks more ambiguous. To this end, we exploit the Gaussian kernel to blur the images. Specifically, the smaller the image, the larger the convolution kernel and the more blurred the transformed image is. In this way, the distance transformation function can simulate the images captured by the camera at different distances.
  }
  \item {\textbf{Angle transformation}: Angle transformation converts an image from the perspective plane to the frontal plane. By using different perspective planes, perspective transformation function can simulate the image captured by a camera at different angles.
  }
  \item {\textbf{Illumination transformation}: The illumination transformation function randomly adjusts the brightness and contrast of the image to simulate the image captured by a camera at different illumination conditions.
  }
  \item{\textbf{Photographing transformation}: Since the input of the object detector is the image captured by a camera, the photographing transformation function adds gaussian noises to the digital image to simulate the image captured by a camera.
  }
\end{itemize}

\begin{figure}[!t]
  \centering
    \includegraphics[width=3.0in]{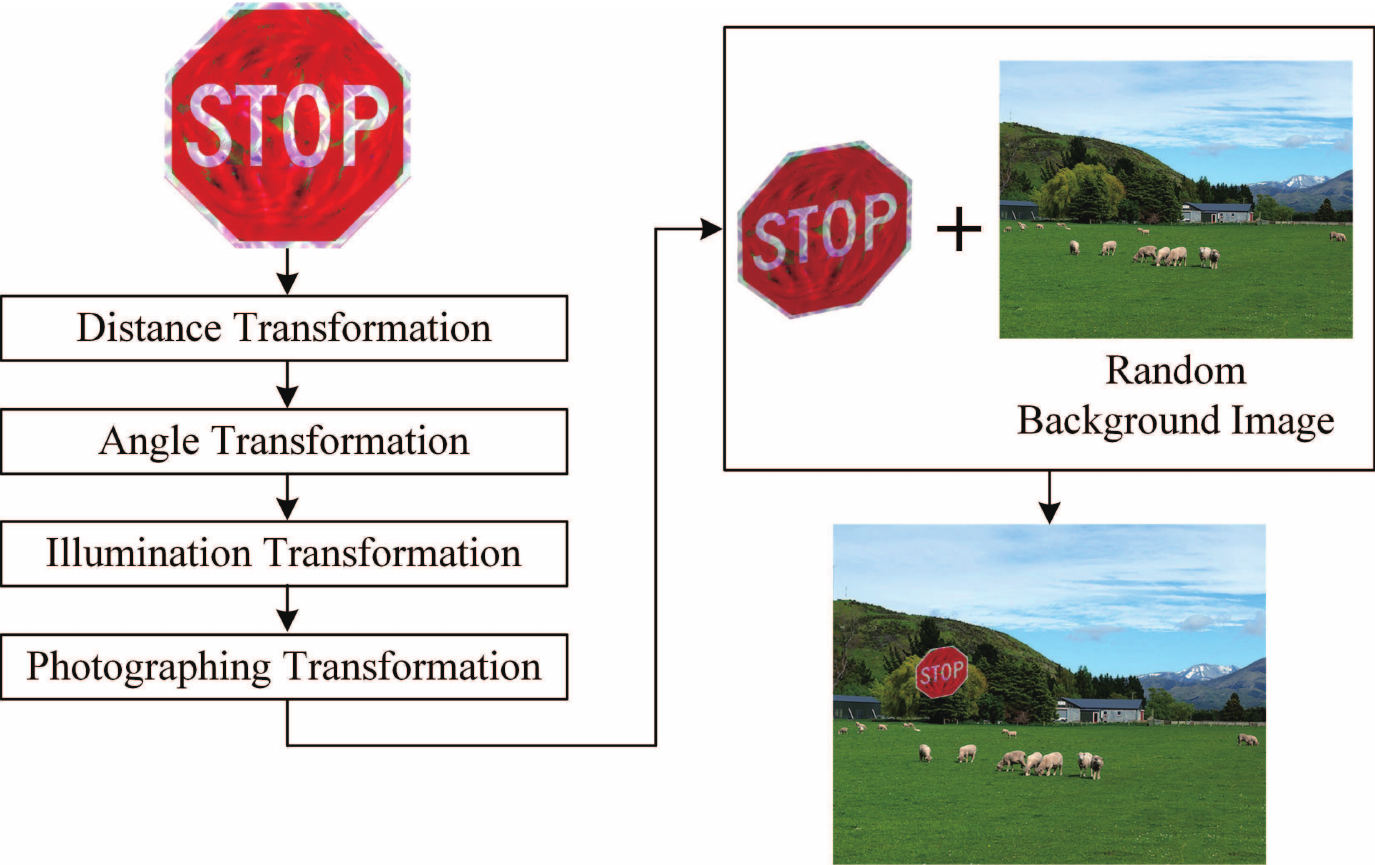}\\
  \caption{Process of image transformations.}
  \label{fig:image_transformation}
\end{figure}

The process of image transformations is shown in Fig. \ref{fig:image_transformation}.
In the optimization of each iteration, first, the generated perturbations are added to the original image.
Then, the above four transformation functions are applied to the image, namely $t(x+\delta)$.
In real physical world, the target object (\textit{e.g.}, the ``STOP'' sign) can appear at various physical scenes that have different context information.
Therefore, to ensure the physical robustness of generated adversarial example, the transformed image will be pasted on different background images $x_{bg}$, \textit{i.e.}, $ t(x+\delta) + x_{bg} $, to simulate the physical attack scenes.
Specifically, at each iteration, we randomly select an image from the Microsoft Common Objects in Context (MS-COCO) dataset \cite{lin2014microsoft} as the background to simulate the physical attack in real-world scenes.
Besides, the object may have different sizes, which depends on the distance between it and the object detector, thus the target object should be randomly resized to adapt to the distance changes.
Finally, the random background containing the adversarial example is used as the input of target object detectors for iterative optimization.

Note that, these image transformations and background images are only used to optimize the generated adversarial example in the generation process, and the final generated physical adversarial example is the original image (\textit{e.g.} a stop sign) with optimized perturbations, which does not contain the background image and these image transformations.
After obtaining the output of the object detector, the proposed method maximizes the difference between the prediction of each object and the true label.
The difference is calculated as follows \cite{chen2018shapeshifter}:
\begin{equation}
  {L_{f}}(x + \delta ,y) = {{\mathbb{E}}_{t\sim T}}\left[ {{1 \over n}\sum\limits_{{r_i} \in rpn(x + \delta )} {J_{f}(t({x_{{r_i}}} + \delta ) + {x_{bg}},y)} } \right]
\end{equation}
where $t(\cdot)$ is a transformation function, and $T$ is the distribution of the transformation function $t$.
Finally, the goal of the iterative optimization is to minimize the following object function \cite{chen2018shapeshifter}:
\begin{equation}
  \mathop {\arg \min }\limits_\delta \alpha {\left\| \delta  \right\|_p} - L_{f}(x+\delta,y)
\end{equation}

\subsection{Constraining the perturbations added on the adversarial examples} \label{sec:proposed_method_2}

The perturbations generated by existing methods can fool object detectors, but these perturbations are unnatural and conspicuous. Humans can easily perceive the anomalies in the generated adversarial examples.
In order to generate adversarial examples that are as similar as possible to the original image, we propose the following two methods to constrain the added perturbations. The first method uses an adaptive mask to constrain the area and intensity of the added perturbations. The second method uses real-world noises to make the added perturbations be more natural. These two methods are introduced in Section \ref{sec:mask} and Section \ref{sec:range}, respectively.

\subsubsection{Constraining the area and intensity of perturbations with an adaptive mask}
\label{sec:mask}

The smaller the area of perturbations added to the original image, the more natural the generated adversarial examples are.
To add the perturbations on the target object only (road sign in this paper), the proposed method uses an adaptive mask to constrain the area where the perturbations are added.
The mask is represented as a matrix $M$, whose dimensions are the same as the shape of the road sign.
Further, to reduce the intensity of adversarial perturbations, the value of each pixel in the mask $M$ is adaptively changed during the optimization process.
Specifically, for each position $(i, j)$, the pixel value of the position on mask $M$ will be multiplied with the adversarial perturbation at the same position.
In other words, if the pixel value in the mask is 0, no perturbation is added to the corresponding position of the original image. If the pixel value in the mask is 1, the perturbation is added to the corresponding position of the original image directly. If the pixel value in the mask is between 0 and 1, the perturbation multiplied by this pixel value is added to the corresponding position of the original image.
Formally, the perturbations added to the original image are constrained as follows:
\begin{equation} \label{fixed_adaptive}
  I_{i,j} + \delta_{i,j} \cdot M_{i,j} = A_{i,j}
\end{equation}
where $I_{i,j}$, $\delta_{i,j}$, $M_{i,j}$, and $A_{i,j}$ are the values of the $i$th row and the $j$th column on the original image ($I$), perturbations ($\delta$), mask ($M$), and the adversarial image ($A$), respectively.

The area of the added perturbations can be represented as ${\left\| {M } \right\|_p}$. If the value of ${\left\| {M} \right\|_p}$ is larger, the area of the added perturbations is larger. Otherwise, the area of the added perturbations is smaller. The final goal of the proposed method is to make the size of the added perturbations as small as possible, the objective function can be further modified as follows:
\begin{equation} \label{adaptive_mask}
  \mathop {\arg \min }\limits_{\delta ,M} {\rm{  }}\alpha {\left\| {M \cdot \delta } \right\|_p} + \beta {\left\| M \right\|_p} - {L_{f}}(x + M \cdot \delta ,y)
\end{equation}
where $ \beta$ is a hyper-parameter for adjusting the weight of ${\left\| M \right\|_p}$.

Compared with the fixed mask that used in existing works \cite{song2018physical,zhao2018practical}, our proposed adaptive mask can constrain the area and intensities of generated adversarial perturbations.
For an fixed mask, it will not be optimized during the whole process of the adversarial example generation, \textit{i.e.}, each pixel value in it remains unchanged \cite{song2018physical,zhao2018practical}.
As a result, according to the Eq. \ref{fixed_adaptive}, the intensities of these perturbations added on a target object (such as stop sign) cannot be constrained.
However, for the adaptive mask in this paper, its pixel values are dynamically changed with these generated perturbations at each iterative optimization. In other words, when generating an adversarial example, we optimize the adversarial perturbations and the mask at the same time (as shown in objective function Eq.(\ref{adaptive_mask}). Therefore, the generated perturbations on our adversarial examples are inconspicuous and can hardly been noticed.

\subsubsection{Generating natural perturbations with real-world noises}
\label{sec:range}
\begin{figure}[htbp]
  \centering
    \includegraphics[width=3.3in]{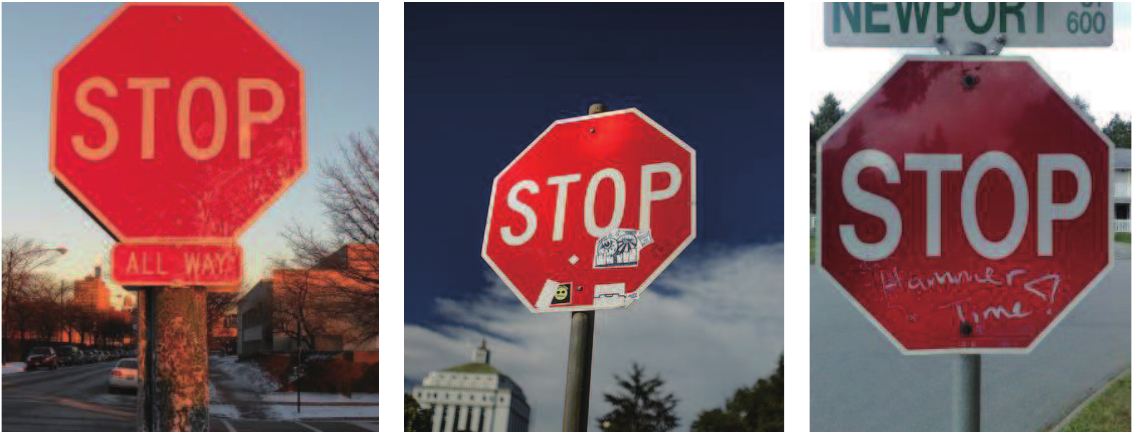}\\
  \caption{Three examples of stop sign in MS-COCO dataset \cite{lin2014microsoft}.}
  \label{fig:real-life_image_example}
\end{figure}

In the real world, there are many real-world images that contain noises. For instance, Fig. \ref{fig:real-life_image_example} shows three examples of stop sign in the MS-COCO dataset \cite{ lin2014microsoft}. It is shown that these real stop signs contain some real-world noises, such as graffiti and stickers. Since humans have no trouble in understanding these images with real-world noises, we aim at generating perturbations that are as similar as possible to the real-world noises. In this way, the generated adversarial examples are natural and similar to the real-world images.

We select real-world images from the MS-COCO dataset \cite{ lin2014microsoft} to extract real-world noises. These extracted real-world noises are denoted as a real-world noise set $P_{real}$. When generating adversarial examples, the proposed method selects the perturbations which are most similar to the noises in the real-world noise set $P_{real}$. Therefore, we define a metric named real-world perturbation score ($RPS$) to represent the similarity between the generated perturbation and the real-world noise, which is calculated by:
\begin{equation}
  RPS(\delta ) = \sum\limits_{p' \in \delta } {\mathop {\min }\limits_{p \in {P_{real}}} {{\left\| {p' - p} \right\|}_2}}
\end{equation}
where $\delta$ is the perturbation vector. $p'$ represents the pixel value in the perturbation vector $\delta$, and $p$ represents the pixel value of the point in the extracted real-world noise set $P_{real}$. If the generated perturbation is similar to the real-world noise, the value of the $RPS$ is small. Otherwise, the value of the $RPS$ is large. Thus, the optimization also needs to minimize the value of the $RPS$.
Finally, after using real-world noises to constrain perturbations, the objective function of the proposed method is as follows:
\begin{equation} \label{final_optimal}
    \mathop {\arg \min }\limits_{\delta ,M} \alpha {\left\| {M \cdot \delta } \right\|_p} + \beta {\left\| M \right\|_p} + \gamma RPS(M \cdot \delta ) - {L_{f}}(x + M \cdot \delta ,y)
\end{equation}
where $\gamma$ is a hyper-parameter that controls the weight of $RPS$ in the optimization process.

The process of the mask generation is described as follows.
First, the mask $M$ is initialized with 0 and 1 (as discussed in Section \ref{sec:exp_setup}.)
Then, to constrain the intensity of the perturbations, the mask $M$ will be multiplied with the adversarial perturbations $\delta$, and the calculated result $M \cdot \delta$ is referred as the ``constrained perturbations''.
Next, the constrained perturbations $M \cdot \delta$ will be added on the target object at each iteration of optimization.
Specifically, the mask $M$ and the adversarial perturbations $\delta$ will be optimized according to three constraint items: $\left\| {M \cdot \delta } \right\|_p$, $RPS(M \cdot \delta)$, and $-{L_{f}}(x + M \cdot \delta ,y)$.
In this process, the pixel values in mask $M$ will be dynamically changed to minimize the sum of the above three items.
Finally, when the value of the objective function is less than the predefined threshold, the adversarial perturbations $\delta$ and the mask $M$ will be generated, respectively.
Note that, the adversarial perturbations should be added on the target object only.
To this end, in our objective function, we use the constraint item $\left\| {M }\right\|_p$ to limit the area of added perturbations.
Therefore, the pixel values on the mask $M$ are also constrained by the item $\left\| {M }\right\|_p$ during the whole optimization process.

This paper focus on the object detection task, and generates the natural and robust adversarial road signs to attack the object detectors in physical world.
However, the proposed method can be applied to other kinds of objects in computer vision domain or even other domains, e.g., network security fields.

The process of the proposed physical adversarial attack method on other objects in computer vision domain are as follows.
First, those image transformation techniques are performed on the generated adversarial example, to ensure its robustness under various physical constraints.
Second, for naturalness, the attacker can capture some background pictures, which contains those physical scenarios that the target object often appears.
The attacker then extracts real-world noises from these pictures to form the new real-world noise set $P^{*}_{real}$, and calculates the $RPS$ during the optimization process.
Third, different target objects may have different shapes, thus the shape of mask requires to be adjusted accordingly, to constrain the area and intensity of added perturbations on a specific target object.
Through the above three steps, the attacker can apply the proposed method to attack other different objects, and generate the natural and robust physical adversarial example in real world.

Besides, the idea behind our proposed method can even be applied to the adversarial example attacks in network security field \cite{Dongqi2020pratical, PierazziPCC20, ApruzzeseCFM19, Chernikova2019Adversarial}.
For example, to attack the network intrusion detectors, the attackers can extract the features and distributions of the normal data points, and then carefully modify these extracted features by adding the adversarial perturbations through the optimization process.
In this way, the attacker can construct a ``natural'' adversarial example, which is extremely similar to a clean data and has the same distribution of these normal data points. As a result, the generated adversarial example will be more difficult to detect, and the adversarial example attacks can be launched in real physical world more covertly.
However, under different attack scenarios, the proposed method requires to be adjusted and modified according to the specific detectors. This requires more theoretical exploration and future works.

\section{Experimental results}\label{sec:experiment}

In this section, we evaluate the success rate of the generated adversarial examples on fooling object detectors in the physical conditions. First, the experimental setup is presented in Section \ref{sec:exp_setup}.
Then, the experimental results of physical attacks are presented in Section \ref{sec:exp_res}.
Third, in Section \ref{sec:com}, the generated adversarial examples are compared with the adversarial examples that generated by existing works.
Finally, the transferability of the adversarial examples generated by the proposed method is evaluated in Section \ref{sec:exp_transf}.

\subsection{Experimental setup}\label{sec:exp_setup}

We use the Faster R-CNN Inception v2 \cite{ren2016faster} and YOLO v2 \cite{redmon2017yolo9000} as the target model of the adversarial attack, respectively.
The pre-trained Faster R-CNN Inception v2 model is available from \cite{Tensorflow_detection_model_zoo}, and the pre-trained YOLO v2 model is available in \cite{yolo}.
In our experiment, the clean stop sign is used as the initial image, and its adversarial examples are generated with the proposed method that discussed in Section \ref{sec:proposed_method}.
We use the Canon printer to print the generated adversarial examples on photographic paper with A3 size. Then, the Nikon D3000 camera with an AF-S 18-55mm lens is used to capture the printed adversarial examples. Finally, the captured image is submitted to the target object detector to evaluate the success rate of generated adversarial examples.

\begin{figure}[htbp]
  \centering
    \includegraphics[width=3.2in]{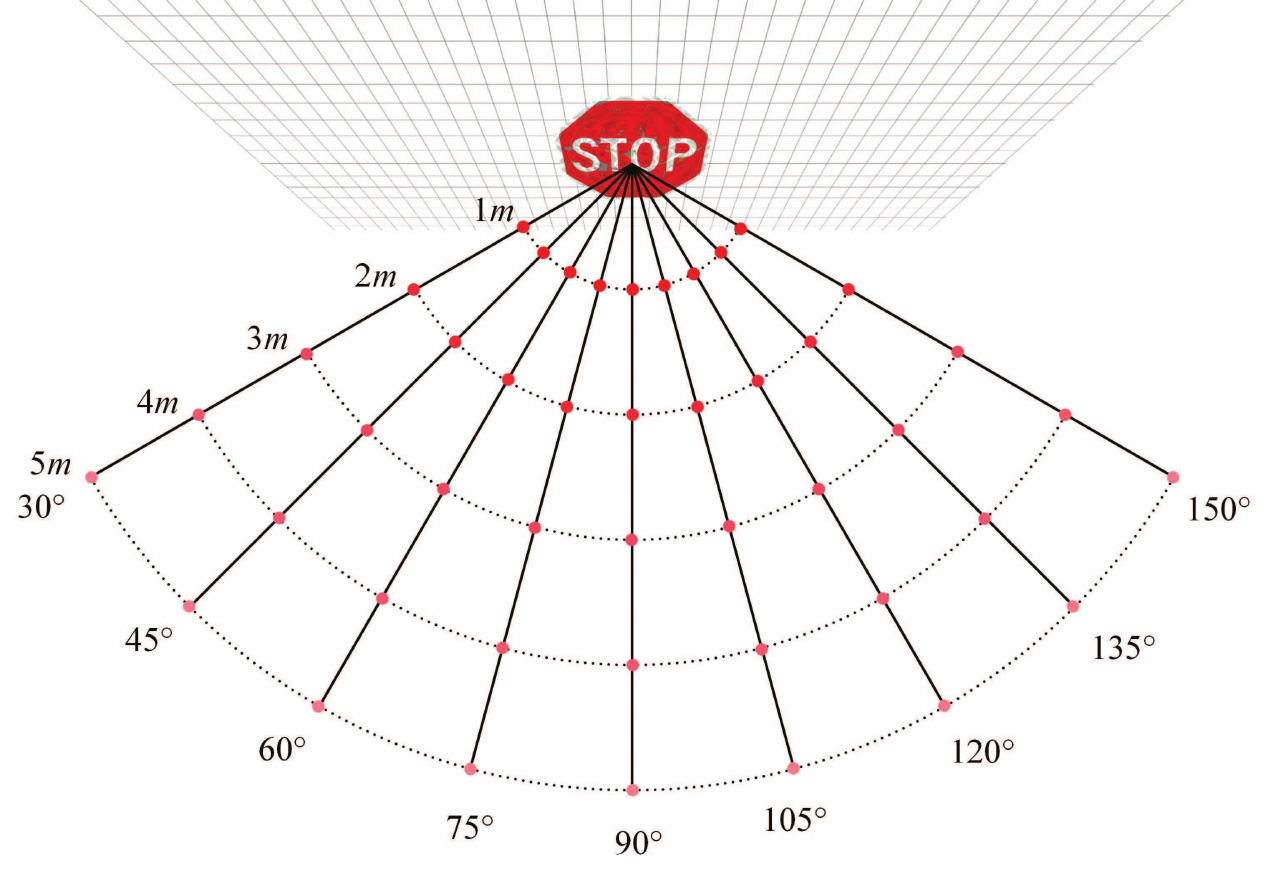}\\
  \caption{Different settings of the photographic distances and angles between the camera and the printed image.}
  \label{fig:exp_setup}
\end{figure}

\textbf{Initialization of adaptive mask $\bm{M}$ and added perturbation $\bm{\delta}$.} The initialization of the adaptive mask and the added perturbations are as follows.

\textbf{- Adaptive mask.} We use an image with the shape of octagon (the same shape as the stop sign) to initialize the mask $M$. The position on the mask corresponds to the position on a stop sign image.
In our experiments, we initialize the adaptive mask with 0 and 1. The reason is as follows.
The proposed method aims to generate the natural adversarial stop sign, while the perturbations added on these white areas (four letters and eight edge regions) are the easiest to be perceived.
Therefore, we should avoid adding the perturbations on these white areas, and make most of the modifications
on the red background region.
To this end, for those positions of four letters (\textit{i.e.}, ``S'', ``T'', ``O'', ``P'') and eight edge regions (\textit{i.e.}, the white stripes), the pixel values on the mask are set to be 0.
The pixel values of the remaining positions (\textit{i.e.}, the red background region) on the mask are all set to be 1.
In this way, even all the pixel values on mask $M$ are updated during the optimization process, the pixel values of these positions that corresponds to the white areas remain small due to such initialization settings.
As a result, when the mask $M$ is multiplied with the adversarial perturbations $\delta$ (\textit{i.e.}, the Eq. (\ref{fixed_adaptive})), the constrained perturbations $M \cdot \delta$ added on the white areas of original stop sign will be small, which ensures the visual naturalness of the generated adversarial example.

\textbf{- Added perturbation.} The added perturbation $\delta$ is initialized by a pure white color image, \textit{i.e.}, the $RGB$ value of each pixel on the perturbation $\delta$ is set to be (255, 255, 255).
These pixel values will be used as the initial perturbations to calculate the real-world perturbation score ($RPS$) and the gradient of the loss function $L_{f}$, respectively.
In our experiments, the adaptive mask $M$ and added perturbation $\delta$ are iteratively optimized by the proposed method with the objective function (Eq. \ref{final_optimal}).

To illustrate the robustness of the generated adversarial examples, we evaluate the success rate of the adversarial examples indoors and outdoors, respectively.
We capture the printed image at different angles and different distances under indoor and outdoor conditions, and submit them to Faster R-CNN Inception v2 detector to detect.
The detailed photographic distance and angle settings are shown in Fig. \ref{fig:exp_setup}. The angle of view between the camera and the stop sign is from $30^\circ$ to $150^\circ$, and the distance between the camera and the stop sign is from $1$ meter to $5$ meters.

\textbf{Evaluation metrics.} We evaluate the generated adversarial examples from the following two aspects: the success rate of adversarial examples and the naturalness of adversarial examples. First, similar to the existing works \cite{eykholt2018robust} and \cite{huang2019upc}, we calculate the success rate $R_s$ of adversarial examples as follows \cite{eykholt2018robust,huang2019upc}:
\begin{equation}
R_s={{\sum\limits_{x \in X,x' \in X'} {\{ f({x_{d,a,i,s}}) = y \wedge f(x{'_{d,a,i,s}}) \ne y\} } } \over {\sum\limits_{x \in X} {\{ f({x_{d,a,i,s}}) = y\} } }}
\end{equation}
where $x_{d,a,i,s}$ represents the original image pasted on the physical scene $s$ captured by the camera at distance $d$, angle $a$, and illumination $i$. Similarly, $x'_{d,a,i,s}$ represents the adversarial example pasted on the physical scene $s$ captured by the camera at distance $d$, angle $a$, and illumination $i$.

Second, we use the size of the added perturbations to evaluate the naturalness of an adversarial example.
If the size of added perturbations is small, the total modifications on the pixel values of an original image (\textit{e.g.}, the stop sign) will be small as well.
As a result, the generated adversarial example is visually similar to the original image and looks natural for humans.
Otherwise, the generated adversarial examples are more conspicuous. In our experiment, we use the standard Euclidean norm ($L_2$-norm) to evaluate the size of the added perturbations, which is calculated as follows \cite{carlini2017adversarial}:
\begin{equation}
\delta {\rm{ = }}{\left\| {{X_{adv}} - X{}_{ori}} \right\|_2}
\end{equation}
where $X_{adv}$ is the generated adversarial stop sign and $X_{ori}$ is the original stop sign.

\subsection{Experimental results of physical adversarial attacks}\label{sec:exp_res}
First, we evaluate the effectiveness of generated adversarial example in real physical world.
The detector is unable to detect the target object whose confidence score is less than the confidence threshold. Specifically, in our experiment, the confidence threshold of Faster R-CNN Inception v2 \cite{ren2016faster} and YOLO v2 \cite{redmon2017yolo9000} detector is set to be 0.2 and 0.4, respectively.
The adversarial example attack is considered to be successful if the prediction result of object detector is not a stop sign or undetected.
Note that, if an adversarial example is simultaneously detected as multiple different objects, the object with the maximum confidence level is selected as the detection result.

\textbf{Faster R-CNN Inception v2.} In this work, we have generated two different adversarial examples against the Faster R-CNN Inception v2 detector by using different combinations of hyper-parameters.
For simplicity, we refer them as Proposed-1 and Proposed-2 (as shown in Table \ref{tab:Comparison} in Section \ref{sec:com}), respectively.

We captured the adversarial stop sign in 45 different positions indoors and outdoors, and submit them to the Faster R-CNN Inception v2 detector to detect. Table \ref{tab:det_res_faster} shows the attack success rates of our generated two adversarial examples under indoor and outdoor physical scenes.
It is shown that, the attack success rate of adversarial examples indoors and outdoors is high up to 73.33\% and 80.00\%, respectively.
The attack success rate of adversarial example Proposed-2 indoors (57.78\%) is lower than outdoors (80.00\%). This is because, the added perturbations on Proposed-2 is small and the light indoors is darker, which causes the camera cannot capture those tiny perturbations under the dark physical scene.
Note that, during the generation of the Proposed-1, the parameter of illumination transformation is set to be small, which makes the generated Proposed-1 more adapted to the dark conditions. As a result, the attack performance of Proposed-1 on Faster R-CNN Inception v2 indoors is better than outdoors.
Overall, the success rate of our proposed physical adversarial example attack is robust, and can successfully cause the Faster R-CNN Inception v2 to make incorrect predictions.

\begin{table}[htbp]
  \centering
  \small
  \caption{Attack success rate of generated stop signs against the Faster R-CNN Inception v2 and YOLO v2 detectors under indoor \& outdoor physical scenes.}
    \begin{tabular}{|c|c|c|c|}
    \hline
    \multicolumn{1}{|c|}{\multirow{2}[3]{*}{\tabincell{c}{Generated \\Stop Signs}}} & \multirow{2}[3]{*}{\tabincell{c}{Target Objector}} & \multicolumn{2}{c|}{Attack Success Rate} \\
\cline{3-4}     &     & Indoors & Outdoors \\
    \hline
    Proposed-1 & \tabincell{c}{Faster R-CNN Inception v2} &73.33\% & 60.00\%  \\
    \hline
    Proposed-2 & \tabincell{c}{Faster R-CNN Inception v2} &57.78\% & 80.00\% \\
    \hline
    Proposed-3 & YOLO v2 &71.11\% & 82.22\% \\
    \hline
    \end{tabular}%
  \label{tab:det_res_faster}%
\end{table}%

\textbf{YOLO v2.} Besides, we exploit the proposed method to attack the YOLO v2 object detector. The generated adversarial example targeting at the YOLO v2 model is referred as Proposed-3 (as shown in Table \ref{tab:Comparison} in Section \ref{sec:com}). The attack performance of Proposed-3 is presented in Table \ref{tab:det_res_faster}.
The attack success rate of Proposed-3 is 82.22\% outdoors, and 71.11\% indoors, which means the physical attacks can fool the YOLO v2 detector at most of the locations.
Similarly, due to the influence of lighting, the attack performance of proposed-3 outdoors is better than the attack performance indoors.

\begin{figure*}[!htbp]
  \centering
    \includegraphics[width=4.4in]{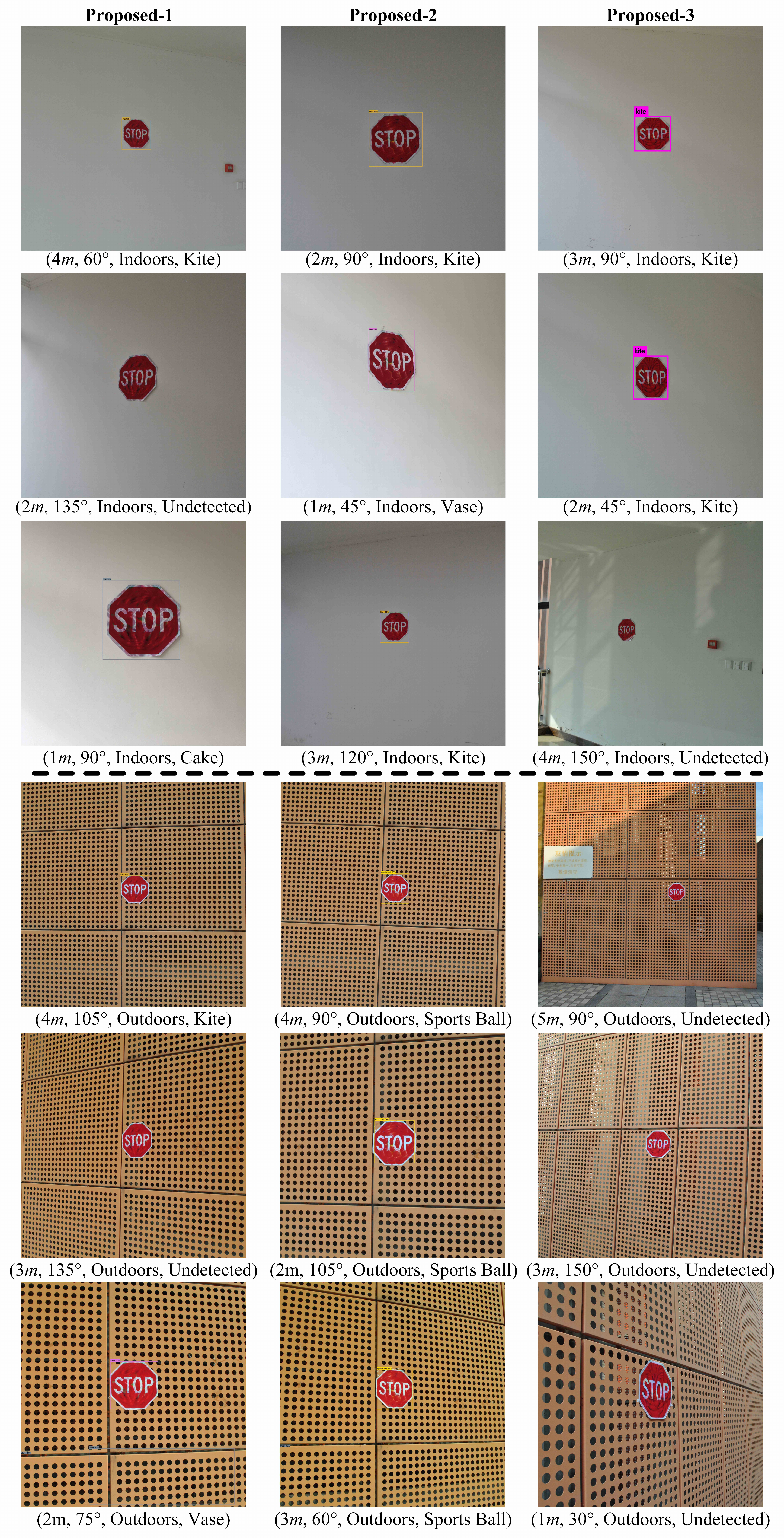}\\
  \caption{Some examples of successful adversarial attacks. The contents in parentheses indicate the distance, angle, indoors/outdoors, and the detection result. The first two columns are the detection results of Faster R-CNN Inception v2 detector, while the third column presents detection results of YOLO v2 detector.}
  \label{fig:exp_of_succ_attacks}
\end{figure*}

Fig. \ref{fig:exp_of_succ_attacks} illustrates some examples of successful adversarial attacks indoors and outdoors.
The contents in parentheses indicate the distance, angle, indoor/outdoor illumination, and the prediction result.
The first two columns are the detection results of Faster R-CNN Inception v2 detector, while the third column shows detection results of YOLO v2 detector.
It is shown that, the printed adversarial stop sign captured at various angles and distances in different illuminations can successfully fool the Faster R-CNN Inception v2 and YOLO v2 detectors, which indicates that the proposed adversarial example attack method is effective under different physical attack scenarios.
Specifically, the generated adversarial stop signs are detected by Faster R-CNN Inception v2 and YOLO v2 as other objects (such as ``Sport Ball'', ``Vase'', ``Kite'', \textit{etc.}) or even undetected.
For example, at the distance of 3m, adversarial stop sign (Proposed-1) captured at $135^\circ$ outdoors can successfully hide itself from the Faster R-CNN Inception v2 detector (\textit{i.e.}, undetected).

To demonstrate the effectiveness of our proposed techniques (image transformation methods, $RPS$ and adaptive mask), we also print the original ``STOP'' image (without any adversarial perturbations).
For fair comparisons, we take the photos for the printed stop sign under the same environment as our generated adversarial examples, to test whether the original stop sign can fool the Faster R-CNN Inception v2 and YOLO v2 detectors.
The experimental results are shown in Table \ref{tab:Comparison}. It is shown that, without the proposed method, the attack success rate of original ``STOP'' sign is 0\% (0/45) indoors and 0\% (0/45) outdoors on two object detection models, which means the physical attacks completely failed.
Therefore, these adversarial transformations performed on the original image are necessary.

\begin{figure}[!htbp]
  \centering
    \includegraphics[width=3.5in]{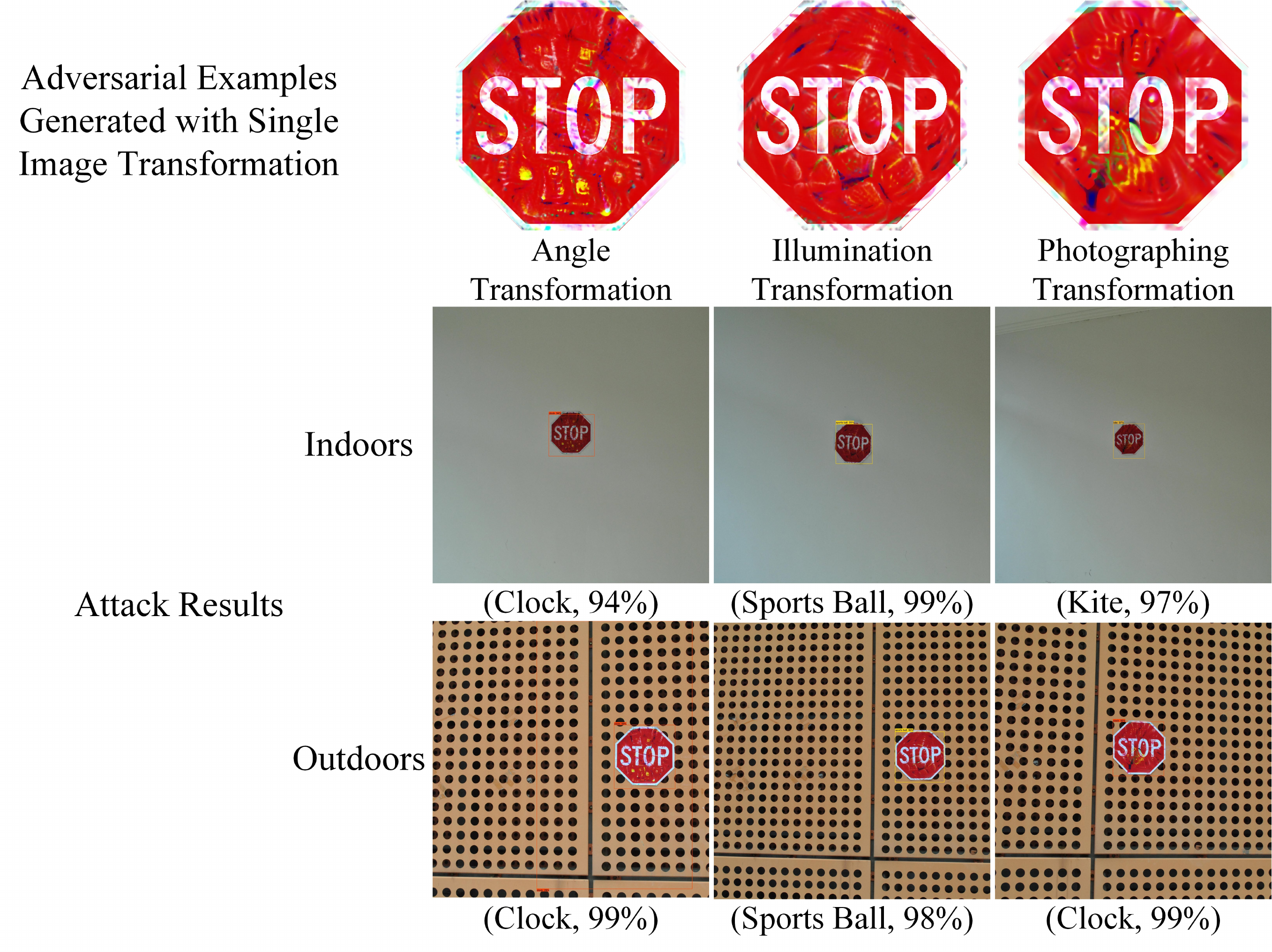}\\
  \caption{Successful attack examples of adversarial stop signs that generated with single image transformation method.}
  \label{fig:ablation}
\end{figure}

Further, we demonstrate whether these processing techniques proposed in Section \ref{sec:proposed_method_1} are effective or not. At each time, we exploit one of these image transformation methods, and remove the others to generate the adversarial example.
Note that, the change in distance is the most common physical constraint when launches the adversarial example attacks in real world,
and many existing works \cite{eykholt2018robust, song2018physical, zhao2018practical} have demonstrated the effectiveness of distance transformation.
Therefore, in this paper, we do not evaluate the distance transformation method.
Specifically, we mainly focus on three different techniques when evaluate the effectiveness of image transformations: angle transformation, photographing transformation and illumination transformation.
We print these adversarial stop signs on photographic papers (A4 size) with the Canon printer.
The generated adversarial examples and some of successful physical attack results are shown in Fig. \ref{fig:ablation}.
It is shown that, even these adversarial stop signs are only performed with a single image transformation, they are still robust under indoor and outdoor physical scenes.
Compared to the original clean stop sign, these adversarial examples are incorrectly detected as ``Clock'', ``Sports Ball'' and ``Kite'', which demonstrates the image transformation techniques are effective and necessary.
In our experiments, we perform the four transformations (angle, photographing, illumination and distance) on the adversarial example at each iteration of optimization, to simulate various possible physical transformations that an adversarial example may undergo in real world. In this way, the physical robustness of adversarial example is guaranteed, and the attack success rate of our generated adversarial example in real physical world is high up to 82.22\% ourdoors and 73.33\% indoors, respectively.

\subsection{Comparison with other methods}\label{sec:com}

Then, to further illustrate the naturalness and robustness of the generated adversarial examples, we compare the adversarial stop sign generated by the proposed method with that generated by \cite{lu2017adversarial} and \cite{chen2018shapeshifter}, from the following aspects: the size of the added perturbations, the success rate of the adversarial examples indoors and outdoors.

In our experiments, we use the same Canon printer to print the adversarial stop signs that generated by \cite{lu2017adversarial} and \cite{chen2018shapeshifter} on photographic papers in A3 size, and use the same Nikon camera to capture photos under the same experimental conditions.
Table \ref{tab:Comparison} shows the comparison results of the adversarial examples generated by the proposed method with that generated by \cite{lu2017adversarial} and  \cite{chen2018shapeshifter}.
\cite{lu2017adversarial}-1 and \cite{lu2017adversarial}-2 are generated by the same method proposed in \cite{lu2017adversarial}, but \cite{lu2017adversarial}-2 adds more perturbations than \cite{lu2017adversarial}-1.
Similarly, \cite{chen2018shapeshifter}-1 and \cite{chen2018shapeshifter}-2 are adversarial stop signs generated by the same method proposed in \cite{chen2018shapeshifter}, and the difference is that \cite{chen2018shapeshifter}-2 adds more perturbations than \cite{chen2018shapeshifter}-1.

\begin{table*}
  \centering
  \small
  \caption{Comparison of the proposed method with existing attack methods}
  \vspace{0.5em}
    \begin{tabular}{|m{5.7em}<{\centering}|m{4.95em}<{\centering}|m{4.95em}<{\centering}|m{4.95em}<{\centering}|m{4.95em}<{\centering}|m{4.95 em}<{\centering}| m{4.95 em}<{\centering}| m{4.95 em}<{\centering}| m{4.95 em}<{\centering}| }
    \hline
     Method &Original image &Proposed-1 & Proposed-2 & Proposed-3 & \cite{lu2017adversarial}-1 & \cite{lu2017adversarial}-2 & \cite{chen2018shapeshifter}-1 &  \cite{chen2018shapeshifter}-2    \\
    \hline
    Generated Adversarial Stop Sign &    \begin{minipage}{0.15\textwidth}
      \includegraphics[width=0.64in]{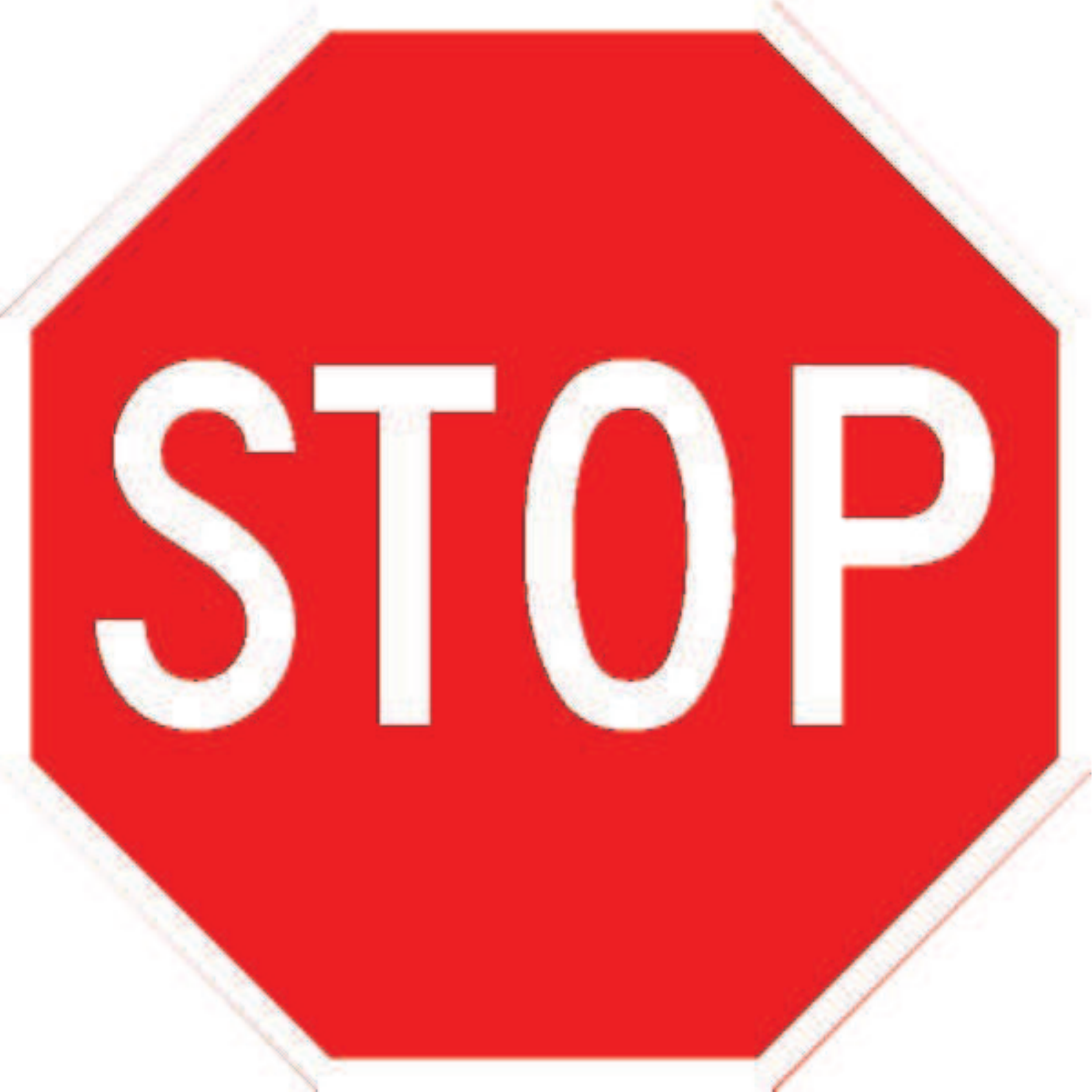}
    \end{minipage}  &
    \begin{minipage}{0.15\textwidth}
      \includegraphics[width=0.65in]{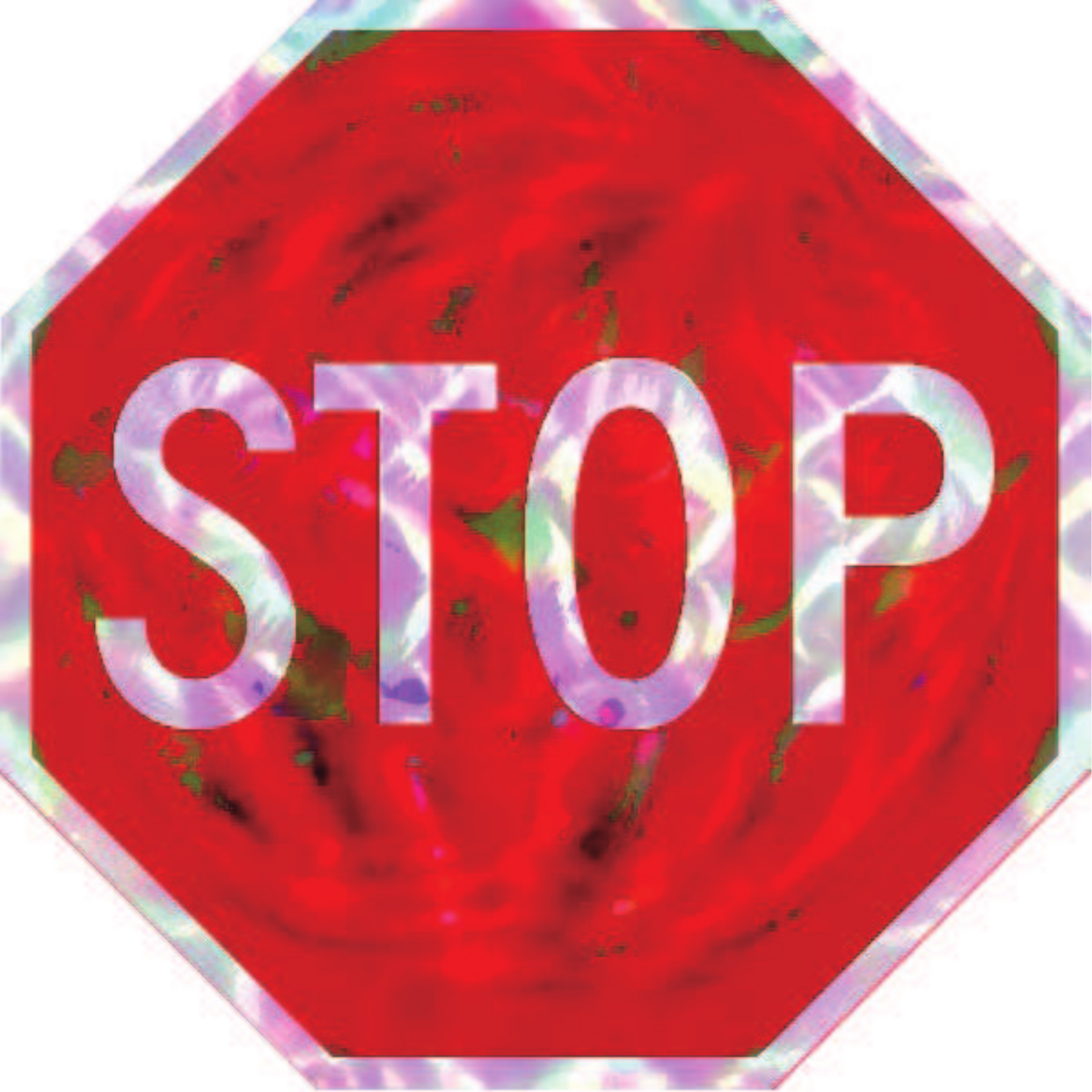}
    \end{minipage}  &   \begin{minipage}{0.15\textwidth}
      \includegraphics[width=0.65in]{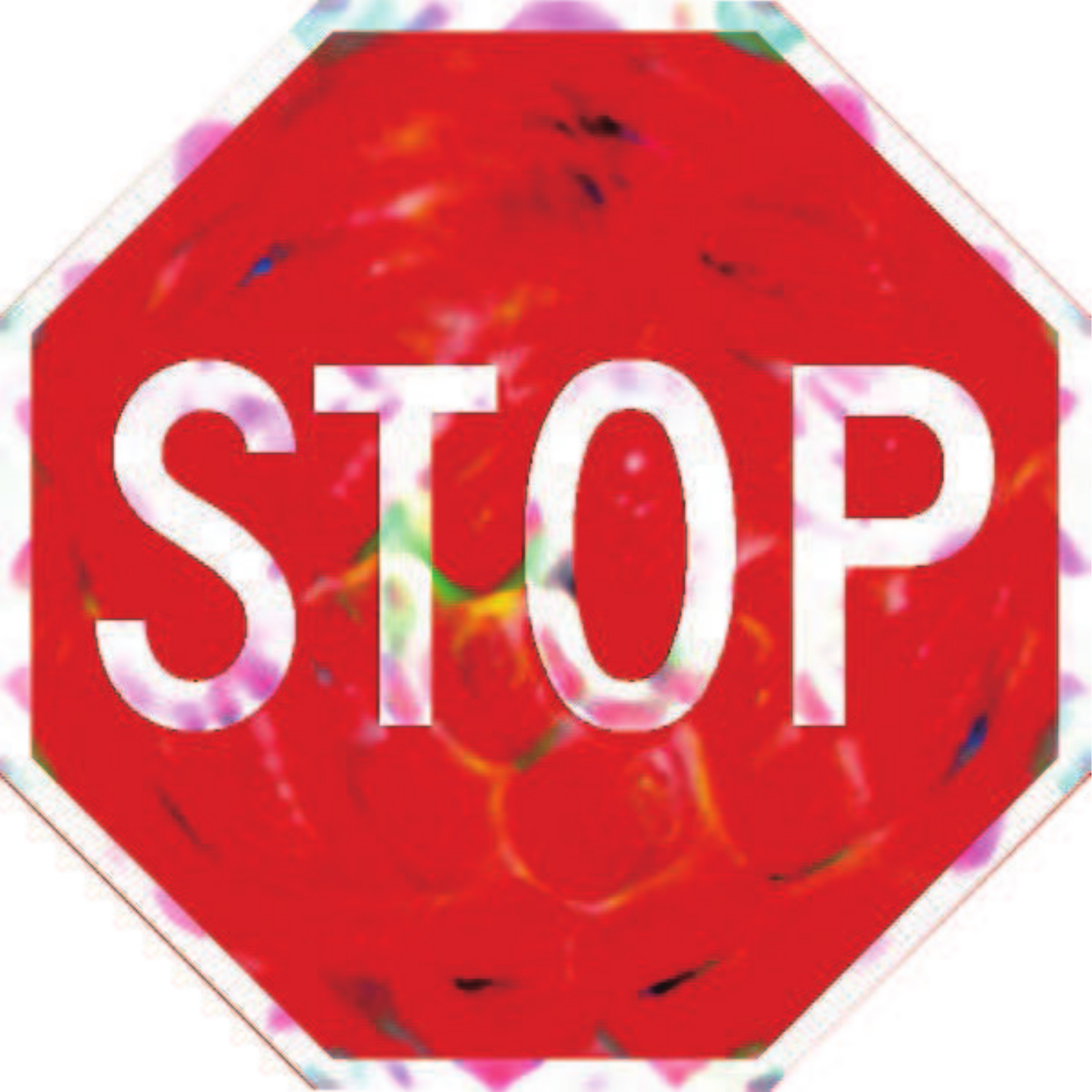}
    \end{minipage} &   \begin{minipage}{0.15\textwidth}
      \includegraphics[width=0.64in]{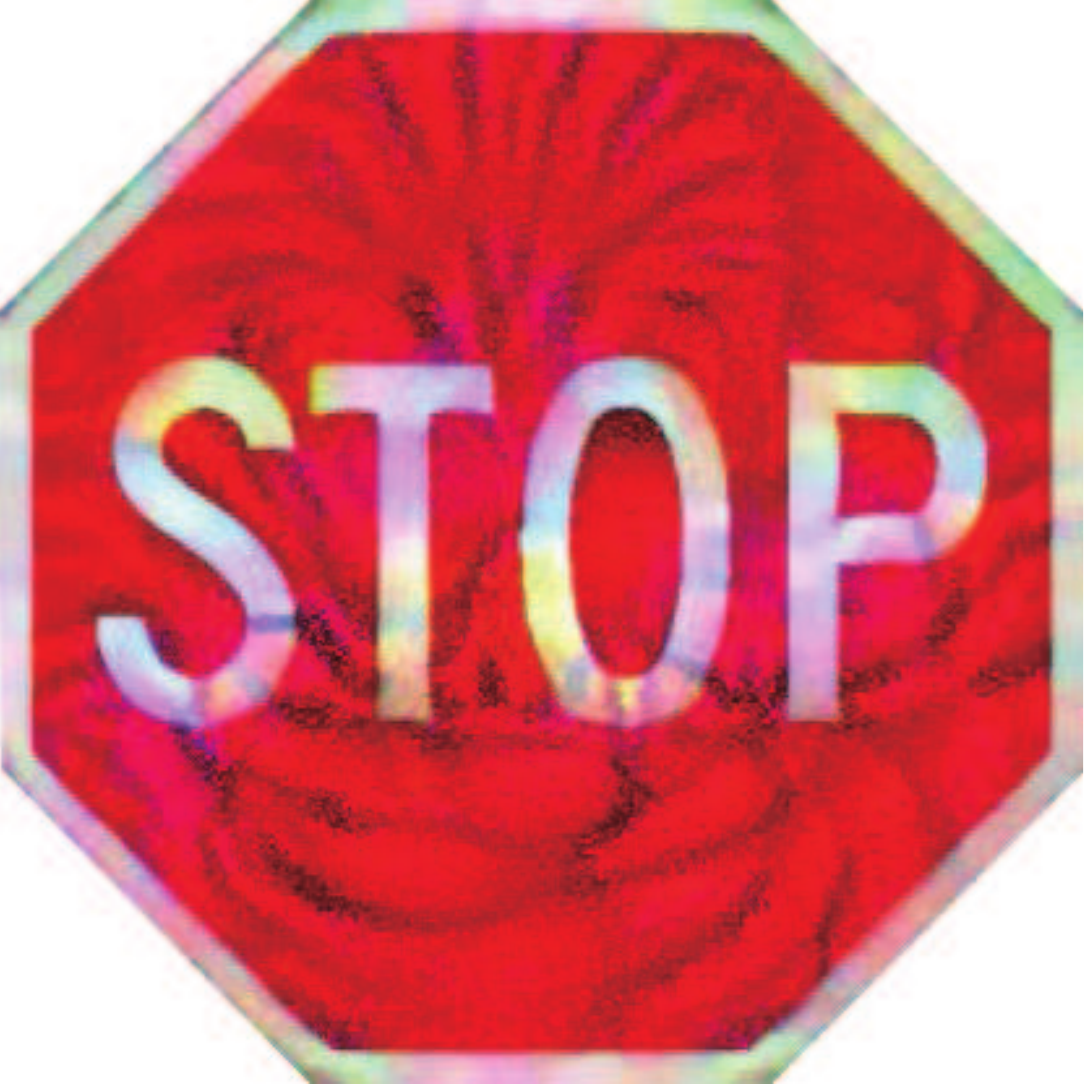}
    \end{minipage}  &
    \begin{minipage}{0.15\textwidth}
      \includegraphics[width=0.65in]{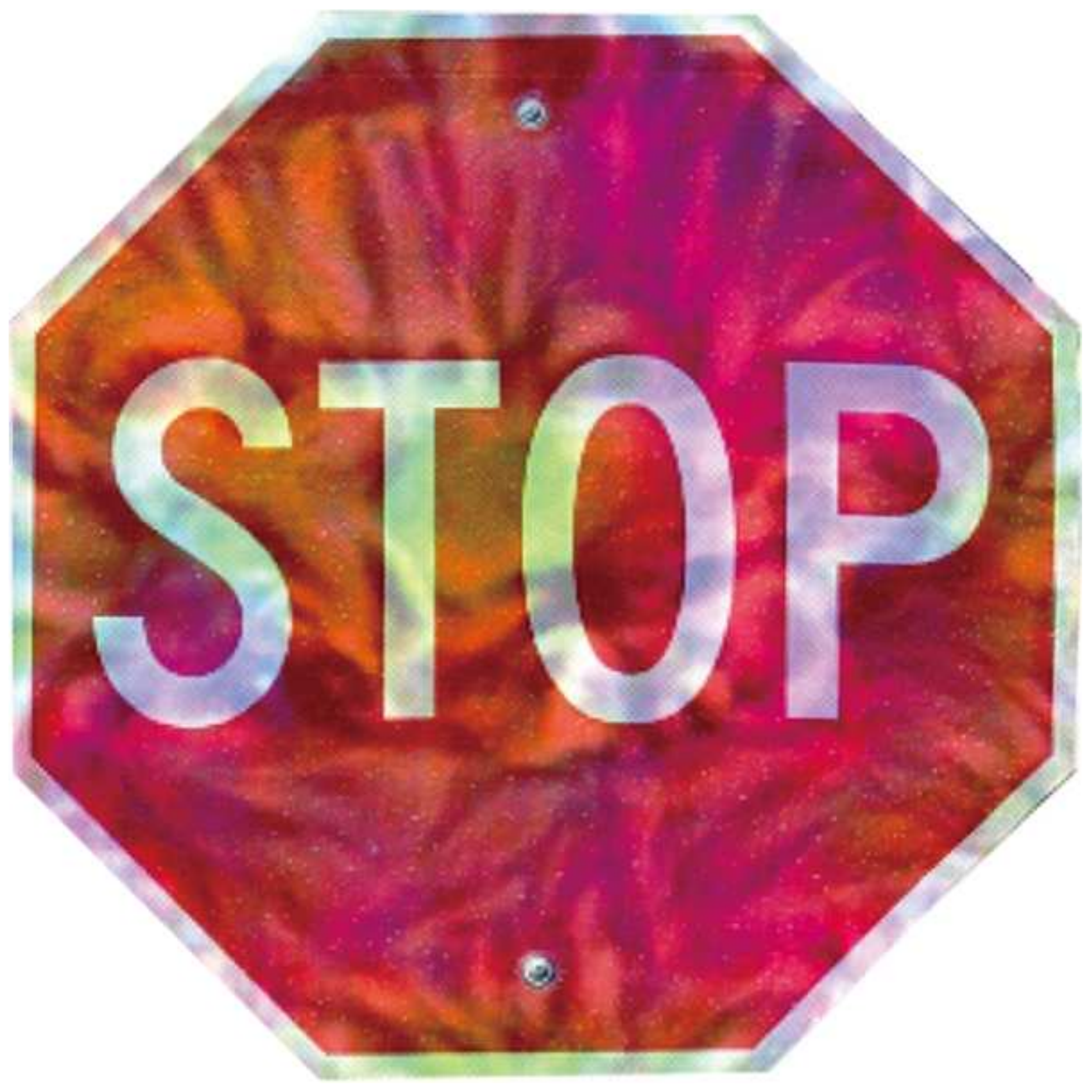}
    \end{minipage}     &   \begin{minipage}{0.15\textwidth}
      \includegraphics[width=0.65in]{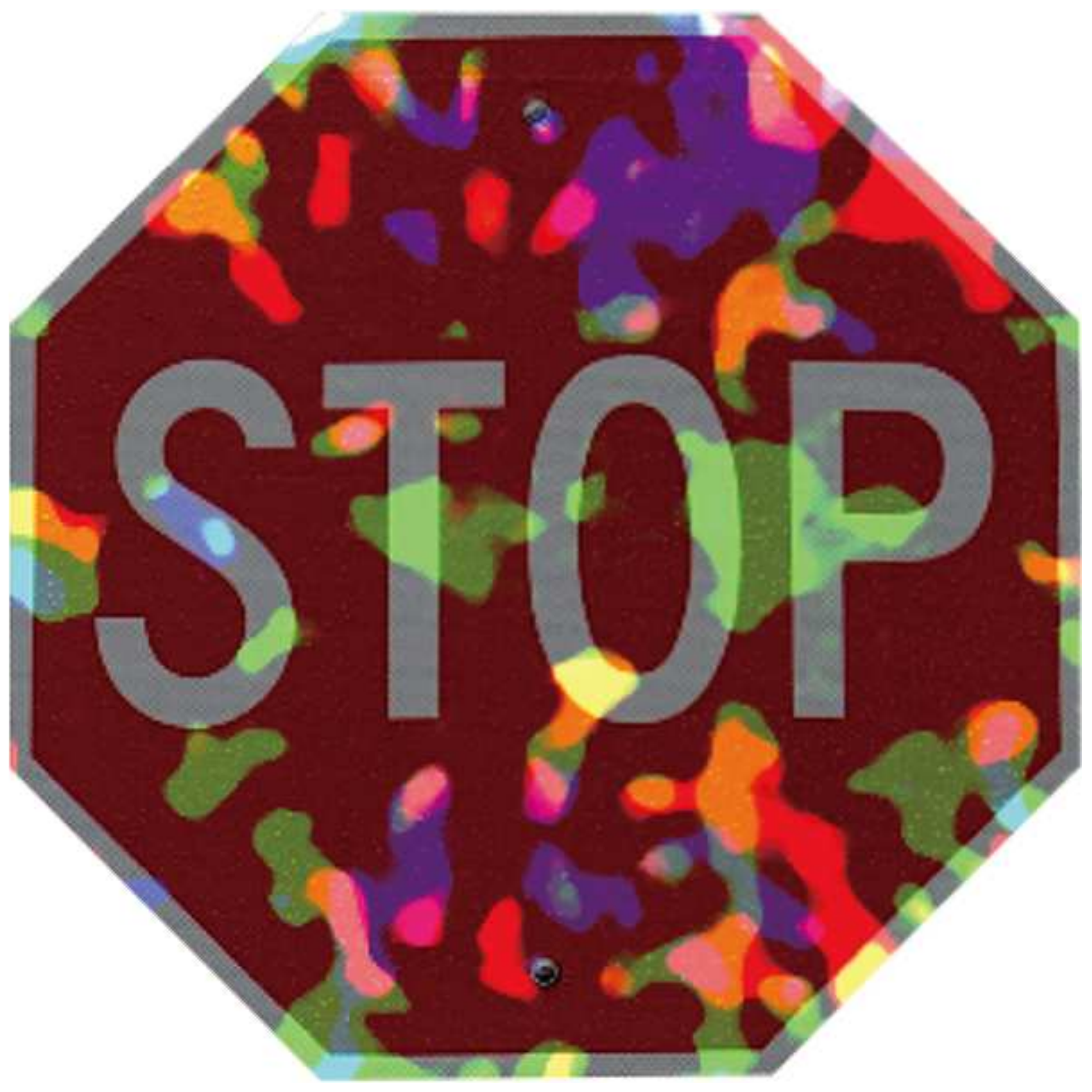}
    \end{minipage}
        &   \begin{minipage}{0.15\textwidth}
      \includegraphics[width=0.65in]{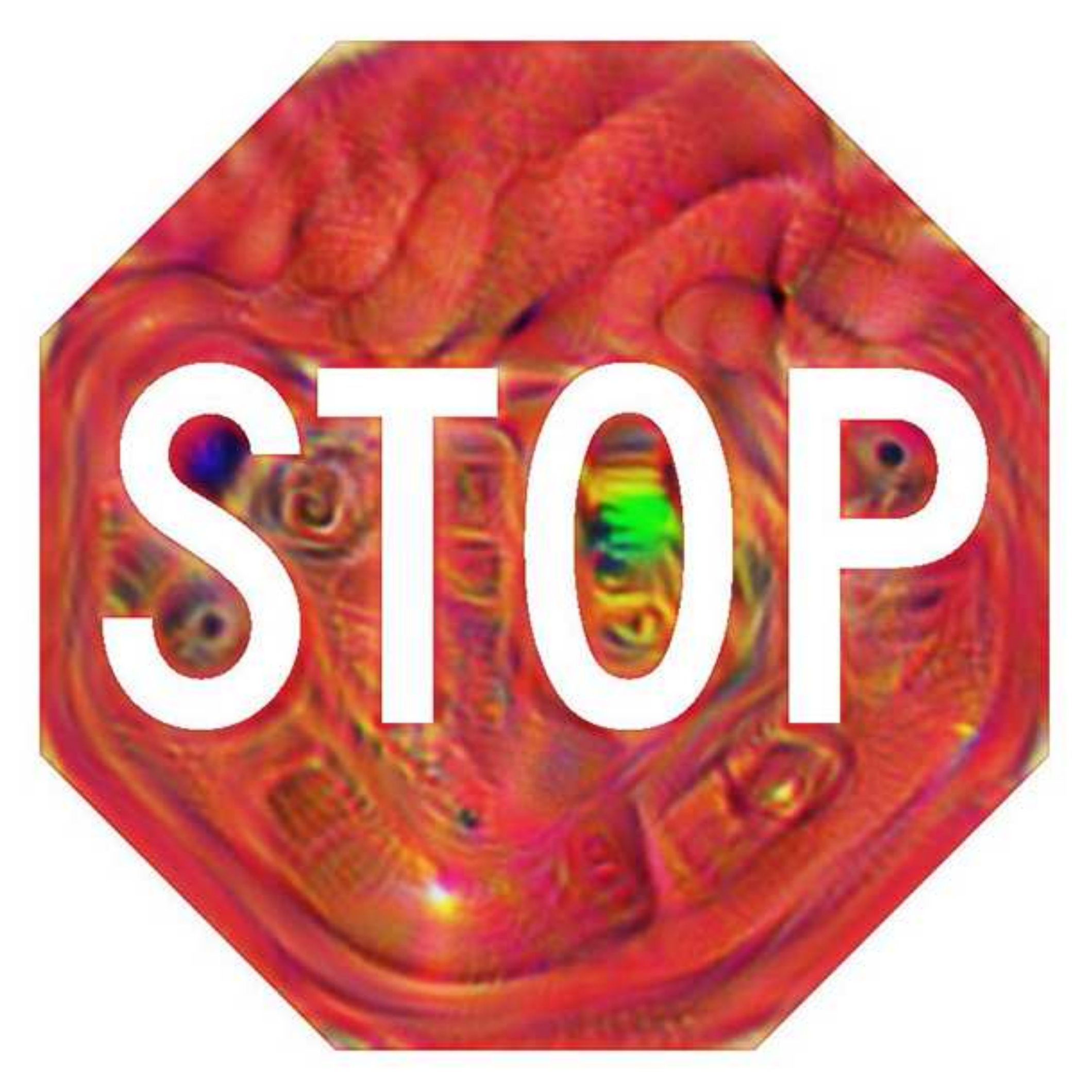}
    \end{minipage}         &   \begin{minipage}{0.15\textwidth}
      \includegraphics[width=0.65in]{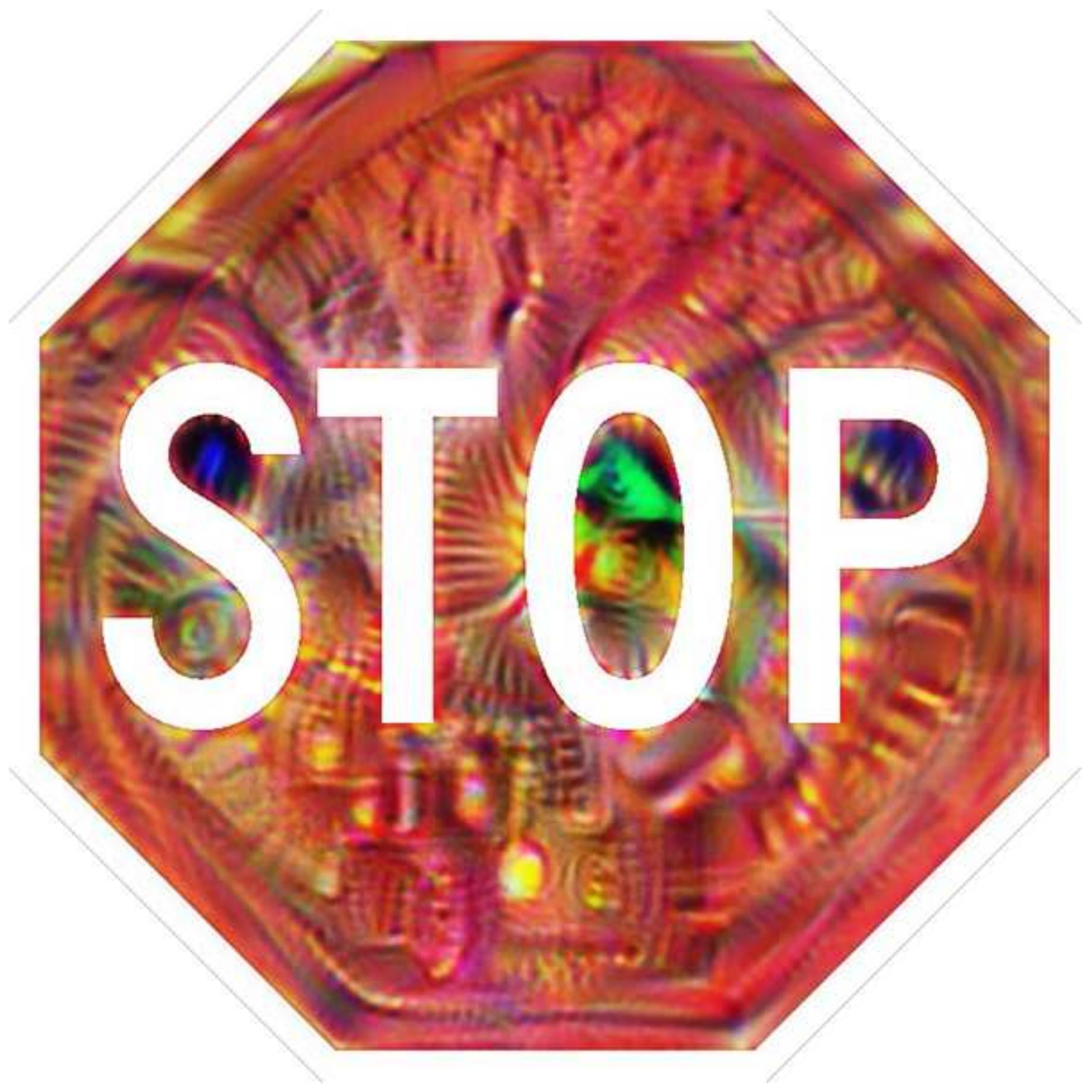}
    \end{minipage}          \\
      \hline
    Size of Added Perturbations &0& 35265.72 & 29361.86 & 39865.59 &65800.86 & 95381.14 & 53989.45 & 65735.61    \\
      \hline
    Success Rate (Indoors) &0.00\%& 73.33\% & 57.78\%& 71.11\% &17.78\% & 75.56\% & 42.22\% & 80.00\%   \\
      \hline
    Success Rate (Outdoors) &0.00\%& 60.00\% & 80.00\%& 82.22\% &26.67\% & 66.67\% & 55.56\% & 93.33\%  \\
    \hline
    \end{tabular}

    \label{tab:Comparison}

\end{table*}

\textbf{Size of perturbations.} The perturbations in the adversarial stop sign generated by the proposed method are much smaller than the perturbations in the adversarial stop signs generated by \cite{lu2017adversarial} and \cite{chen2018shapeshifter}.
The sizes of the added perturbations generated by \cite{lu2017adversarial}-1, \cite{lu2017adversarial}-2, \cite{chen2018shapeshifter}-1, and \cite{chen2018shapeshifter}-2 are 65800.86, 95381.14, 53989.45, and 65735.61, respectively.
However, the sizes of the added perturbations on our generated adversarial examples are only 35265.72 (Proposed-1), 29361.86 (Proposed-2) and 39865.59 (Proposed-3), respectively,
which are much smaller than the perturbations in related works.
As shown in Table \ref{tab:Comparison}, compared to the adversarial perturbations generated by \cite{lu2017adversarial, chen2018shapeshifter}, the difference between the our generated perturbations and the original image is more imperceptible, and the generated adversarial stop signs looks more natural.

\begin{figure}[!htbp]
  \centering
    \includegraphics[width=3.3in]{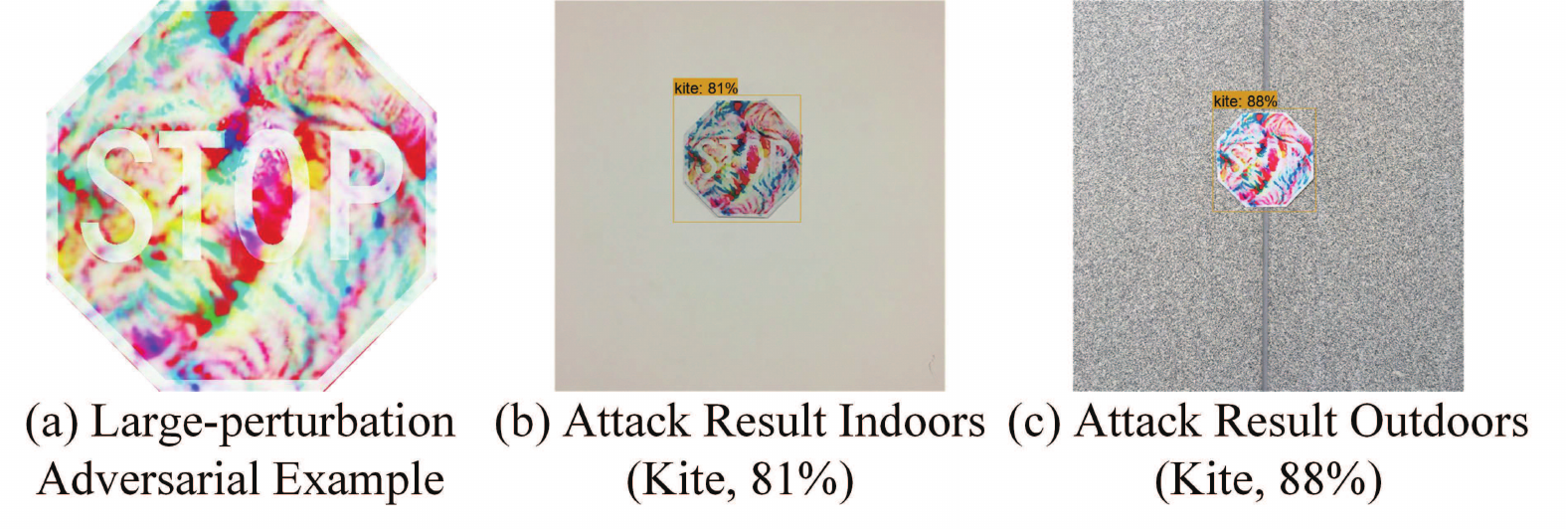}\\
  \caption{Generated large-perturbation adversarial example without the proposed $RPS$ and adaptive mask techniques, as well as the physical attack results indoors and outdoors.}
  \label{fig:example_large}
\end{figure}

In real world, people may use machine learning based techniques to automate the process in computer vision tasks, such as object detection. However, this does not imply that these tasks are out of humans' supervision, especially for those security or safety critical scenarios. For example, when the added perturbations on a ``STOP'' sign is too large and even seriously affect humans' understanding on it, the local supervision department will replace the attacked objects in time to avoid potential safety risks. As a result, the adversarial example attacks in real physical world will fail.
To better illustrate this, Fig. \ref{fig:example_large} shows a large-perturbation adversarial example that generated without the proposed $RPS$ and adaptive mask techniques.
Specifically, we have removed the first three constraint items (\textit{i.e.}, $\alpha {\left\| {M \cdot \delta } \right\|_p}$, $\beta {\left\| M \right\|_p}$, $\gamma RPS(M \cdot \delta )$) in our final objective function. In our proposed method, these three constraint items are used to limit the intensity and area of perturbations, and ensure the naturalness of generated adversarial example.
It is shown that, the large-perturbation adversarial example can successfully fool the object detector, and is incorrectly detected as ``Kite''.
However, the added perturbations on the stop sign is large. This will cause strong visual conflict to humans, and result in the failure of physical adversarial example attacks.

\textbf{Attack success rate indoors and outdoors.} Under indoor conditions, compared with the adversarial stop signs generated by existing works \cite{lu2017adversarial, chen2018shapeshifter}, the attack success rate of our generated adversarial examples (73.33\%) is much higher than the success rate of \cite{lu2017adversarial}-1 (17.78\%) and \cite{chen2018shapeshifter}-1 (42.22\%), and is close to the success rate of \cite{lu2017adversarial}-2 (75.56\%).
Under outdoor conditions, the generated adversarial stop signs can achieve better attack performance than \cite{lu2017adversarial}-1, \cite{lu2017adversarial}-2 and \cite{chen2018shapeshifter}-1. The outdoor attack success rate of Proposed-2 and Proposed-3 is high up to 80.00\% and 82.22\%, respectively, which is significantly higher than the attack success rate of \cite{lu2017adversarial}-1 (26.67\%), \cite{lu2017adversarial}-2 (66.67\%) and \cite{chen2018shapeshifter}-1 (55.56\%) outdoors.

The success rate of our generated adversarial stop sign is slightly lower than the attack success rate of \cite{chen2018shapeshifter}-2 indoors and outdoors.
The reason is that, the work \cite{chen2018shapeshifter} adds much larger perturbations in the adversarial stop sign \cite{chen2018shapeshifter}-2, which makes the object detector more easier to recognize the added perturbations.
However, the adversarial stop sign generated by \cite{chen2018shapeshifter}-2 is visually unnatural and can be easily noticed, while our generated adversarial example is more natural and will not arouse humans' suspicions.
Overall, the generated adversarial examples (Proposed-1, Proposed-2 and Proposed-3) can achieve high attack success rates with less conspicuous perturbations.

\begin{table*}[t]
  \centering
  \small
  \caption{Transfer attack success rate of generated adversarial examples (Proposed-1 and Proposed-2). The Faster R-CNN Inception v2 model is used as the white-box model, and the Faster R-CNN ResNet-50, SSD Inception v2, SSD MobileNet v2 and YOLO v2 models are used as the black-box models.}
    \begin{tabular}{|c|c|c|c|c|c|}
    \hline
    \multicolumn{2}{|c|}{\multirow{3}[4]{*}{Model}} & \multicolumn{4}{c|}{Attack Success Rate} \\
\cline{3-6}    \multicolumn{2}{|c|}{} & \multicolumn{2}{c|}{Indoors} & \multicolumn{2}{c|}{Outdoors} \\
\cline{3-6}    \multicolumn{2}{|c|}{} & Proposed-1 & Proposed-2 & Proposed-1 & Proposed-2 \\
    \hline
    White-box & Faster R-CNN Inception v2 & 73.33\% & 57.78\% & 60.00\% & 80.00\% \\
    \hline
    \multirow{4}[3]{*}{Black-box} & Faster R-CNN ResNet-50 & 15.56\% & 2.22\% & 4.44\% & 4.44\% \\
\cline{2-6}          & SSD Inception v2 & 33.33\% & 20.00\% & 31.11\% & 13.33\% \\
          & SSD MobileNet v2 & 51.11\% & 57.78\% & 44.44\% & 31.11\% \\
\cline{2-6}          & YOLO v2 & 4.44\% & 0.00\% & 6.67\% & 4.44\% \\
    \hline
    \end{tabular}%
  \label{tab:transfer_12}%
\end{table*}%

\subsection{Transferability across different target models}\label{sec:exp_transf}
Finally, to evaluate the proposed method in black-box scenarios, we evaluate the transferability of the generated adversarial examples on different object detectors. We generate adversarial examples based on a white-box model, and evaluate these generated adversarial examples on other black-box models.

This paper has generated the adversarial examples against two different object detectors, Faster R-CNN Inception v2 \cite{ren2016faster} and YOLO v2 \cite{redmon2017yolo9000}.
The Proposed-1 and Proposed-2 are generated targeting Faster R-CNN Inception v2 detector, while the Proposed-3 is generated targeting YOLO v2 detector.
For Proposed-1 and Proposed-2, the Faster R-CNN Inception v2 \cite{ren2016faster} model is used as the white-box model, and the Faster R-CNN ResNet-50 \cite{ren2016faster}, SSD Inception v2 \cite{Liu2016SSD}, SSD MobileNet v2 \cite{Liu2016SSD} and YOLO v2 \cite{redmon2017yolo9000} models are used as black-box models.
For Proposed-3, the YOLO v2 \cite{redmon2017yolo9000} model is regarded as the white-box model, while the Faster R-CNN Inception v2 \cite{ren2016faster}, Faster R-CNN ResNet-50 \cite{ren2016faster}, SSD Inception v2 \cite{Liu2016SSD} and SSD MobileNet v2 \cite{Liu2016SSD} models are used as the black-box models.
In this experiment, we download the pre-trained Faster R-CNN models (Faster R-CNN Inception v2 \& Faster R-CNN ResNet-50) and SSD models (SSD Inception v2 \& SSD MobileNet v2) from \cite{Tensorflow_detection_model_zoo}, and the pre-trained YOLO v2 model is available in \cite{yolo}.

\textbf{Transferability of Proposed-1 and Proposed-2.} Table \ref{tab:transfer_12} shows the transfer attack success rate of two adversarial examples (Proposed-1 and Proposed-2) that generated targeting Faster R-CNN Inception v2 model.
It is shown that, the two generated adversarial examples have good transferability on the SSD Inception v2 \cite{Liu2016SSD} and SSD MobileNet v2 \cite{Liu2016SSD} models.
Under indoor conditions, the success rates of Proposed-1 and Proposed-2 on SSD MobileNet v2 model is 51.11\% and 57.78\%, respectively. Besides, the transfer attack success rate of Proposed-1 on SSD Inception v2 model is 33.33\% indoors and 31.11\% outdoors, respectively.
Compared to the YOLO v2 and Faster R-CNN ResNet-50 model, the transferability of two generated adversarial examples (Proposed-1 and Proposed-2) on two SSD models are much better indoors and outdoors.
This is because, the feature extraction network of SSD model is more complex, and it uses the ``extra feature layers'' \cite{Liu2016SSD} to further extract the features from an input image.
In this way, these added adversarial perturbations (especially these tiny ones) are easier to be recognized by SSD detector.
Therefore, under the indoor and outdoor conditions, the attack performances of our generated adversarial examples on SSD models are higher than the attack performances on YOLO v2 model and Faster R-CNN ResNet-50 model.

It is also shown that, the transfer success rates of two generated adversarial examples on Faster R-CNN ResNet-50 model is a little bit higher than the success rates on YOLO v2.
The reason is as follows.
The working mechanism of YOLO v2 are completely different from the target object detector Faster R-CNN Inception v2. The YOLO v2 is the one-stage object detector \cite{redmon2017yolo9000}, while the Faster R-CNN Inception v2 is the two-stage detector \cite{ren2016faster}.
However, the Faster R-CNN ResNet-50 model has the same detection structure and the same detection process as the Faster R-CNN Inception v2.
As a result, compared to YOLO v2 model, the generated adversarial example can transfer better to the Faster R-CNN ResNet-50 model.
Note that, the Faster R-CNN ResNet-50 and Faster R-CNN Inception v2 use different feature extractors extracting the features from an image, which has a great influence on the transferability of the adversarial examples. Therefore, the transferability of two generated adversarial examples (Proposed-1 and Proposed-2) on Faster R-CNN ResNet-50 is also limited.

\textbf{Transferability of Proposed-3.} Table \ref{tab:transfer_3} presents the transfer attack success rate of adversarial examples (Proposed-3) that generated targeting YOLO v2 model.
It is shown that, similar to Proposed-1 and Proposed-2, the generated adversarial example Proposed-3 transfers well on two SSD models.
Specifically, the transfer attack success rate on SSD MobileNet v2 model is 77.78\% (indoors) and 55.56\% (outdoors), while the indoor attack success rate is even higher than the attack success rate on YOLO v2 model (73.33\%).
Besides, the transfer attack success rate of Proposed-3 on SSD Inception v2 model is 37.78\%.
Note that, since the YOLO v2 and Faster R-CNN models have completely different network structures and working mechanisms, the transfer performance of Proposed-3 on two Faster R-CNN models (Faster R-CNN Inception v2 and Faster R-CNN ResNet-50) is relatively poor.

In conclusion, the adversarial examples generated targeting Faster R-CNN Inception v2 (Proposed-1 \& Proposed-2) and YOLO v2 (Proposed-3) can be successfully transferred to two SSD models (SSD Inception v2 \& SSD MobileNet v2). This demonstrate that, for those models which have different working mechanisms and network structures from the white-box models, the generated adversarial examples are still effective.

\begin{table}[htbp]
  \centering
  \small
  \caption{Transfer attack success rate of generated adversarial example (Proposed-3). The YOLO v2 model is used as the white-box model, and the Faster R-CNN Inception v2, Faster R-CNN ResNet-50, SSD Inception v2 and SSD MobileNet v2 models are used as the black-box models.}
    \begin{tabular}{|m{4.3em}<{\centering}|m{11.5em}<{\centering}|m{3.3em}<{\centering}|m{3.3em}<{\centering}|}
    \hline
    \multicolumn{2}{|c|}{\multirow{2}[4]{*}{Model}} & \multicolumn{2}{c|}{Success Rate} \\
\cline{3-4}    \multicolumn{2}{|c|}{} & Indoors & Outdoors \\
    \hline
    White-box & YOLO v2 &   71.11\%    & 82.22\% \\
    \hline
    \multirow{4}[8]{*}{Black-box} &  Faster R-CNN Inception v2 &    2.22\%   &  0.00\%\\
\cline{2-4}          & Faster R-CNN ResNet-50 &    2.22\%   &  8.89\%\\
\cline{2-4}          & SSD Inception v2 &    35.56\%   &  37.78\%\\
\cline{2-4}          & SSD MobileNet v2 &    77.78\%   &  55.56\%\\
    \hline
    \end{tabular}%
  \label{tab:transfer_3}%
\end{table}%

\section{Conclusion}\label{sec:conclusion}

This paper proposes a natural and robust physical adversarial example attack method for object detectors.
We perform a range of image transformations to simulate different physical conditions during the iterative optimization process, to guarantee the robustness of the generated adversarial example.
Meantime, for visual naturalness, this paper proposes two novel techniques, the adaptive mask and the real-world perturbation score ($RPS$), to constrain the added adversarial perturbations.
Experimental results in various physical conditions show that, our generated adversarial example can achieve a high attack success rate with natural perturbations.
Compared with these existing adversarial example generation methods, the proposed method obtains a good balance between the robustness and the naturalness of the generated adversarial example.
In addition, our generated adversarial examples can transfer from target white-box models to other different models.
This paper reveals a serious threat, \textit{i.e.}, practical physical adversarial example attack, and further demonstrates the possibility of constructing an adversarial example which is both natural and robust in real-world conditions.
Such physical adversarial example attacks will be more difficult to be observed, which highlight the urgency to develop reliable defense techniques.

\section*{Acknowledgement}
This work is supported by the National Natural Science Foundation of China (No. 61602241).

\section*{References}
\bibliographystyle{model3-num-names}
\bibliography{ref}

\end{document}